%% file: acl_latex.tex
\documentclass[11pt]{article}

\usepackage[final]{acl}

\usepackage{times}
\usepackage{latexsym}

\usepackage[T1]{fontenc}

\usepackage[utf8]{inputenc}

\usepackage{microtype}

\usepackage{inconsolata}

\usepackage{graphicx}
\usepackage{amsmath}
\usepackage{amssymb}
\usepackage{booktabs}
\usepackage{multirow}

\usepackage{colortbl}

\title{LoRU: Uncertainty Is Relative in LLM Hallucination Detection}

\author{
  Yuji Zhang\textsuperscript{1,2}\thanks{Equal contribution.}, Rong Xu\textsuperscript{3}\footnotemark[1], Christie Djidjev\textsuperscript{3}, Jiawei Xu\textsuperscript{3}, Jinrui Fang\textsuperscript{3}, \\\textbf{ChengXiang Zhai}\textsuperscript{2}\thanks{Corresponding authors: \texttt{czhai@illinois.edu}, \texttt{ying.ding@austin.utexas.edu}.}, \textbf{Ying Ding}\textsuperscript{3}\footnotemark[2] \\
  \textsuperscript{1}CUNY CSI~~~~
  \textsuperscript{2}UIUC~~~~
  \textsuperscript{3}UT Austin \\
  \texttt{yuji.zhang@csi.cuny.edu}, \texttt{rongxu@utexas.edu}
}

\begin{document}
\maketitle

\begin{abstract}
Uncertainty-based hallucination detection uses uncertainty signals to assess the risk of hallucinations in generated responses. In long, multi-step reasoning, these signals are distributed across tokens and reasoning steps, raising a key question: how should uncertainty evidence from different parts of a generation be combined to distinguish correct from incorrect final answers? We systematically examine two properties relevant to this aggregation problem. We find that preceding context can substantially shift the uncertainty assigned to the same downstream answer, while uncertainty dependence quickly weakens with increasing distance between tokens.
Motivated by the findings, we introduce \textbf{Lo}cal \textbf{R}elative \textbf{U}ncertainty (\textbf{LoRU}), a training-free plug-in framework that uses local uncertainty comparisons to guide evidence aggregation. To account for context-dependent uncertainty shifts, LoRU uses relative ranks rather than absolute magnitudes alone. Guided by the observed decay of uncertainty dependence with distance, it computes these ranks within sliding windows of recent tokens.
Across 65 experimental settings spanning different models, datasets, and uncertainty estimation methods, LoRU's aggregation strategies achieve AUROC improvements in 76.9\% of cases, with an average relative gain of 5.5\%.
\end{abstract}

\input{sections/introduction}

\input{sections/relatedwork}

\input{sections/understanding}

\input{sections/method}

\input{sections/experiments}

\section{Conclusion and Future Work}
\label{sec:conclusion}

We investigate how uncertainty evidence across
reasoning trajectories can be combined to assess
final-answer correctness. We identify two key
properties: context-dependent uncertainty shifts
and distance-dependent decay. Motivated by these
findings, we introduce LoRU, a training-free
framework that uses local relative ranks to guide
uncertainty aggregation. Experiments across diverse
models, tasks, and estimators demonstrate its
effectiveness in hallucination detection.
Future work could extend local relative uncertainty
from detecting unreliable responses to guiding
reasoning itself, identifying when and where
verification, additional evidence, or reasoning
revision is needed.

\section*{Limitations}

The effectiveness of different LoRU aggregation
strategies varies across reasoning tasks, models,
and uncertainty estimators. Our analyses suggest
that these differences relate to how uncertainty
evidence is distributed across a response and how
the ranking signal aligns with the original
uncertainty measure used for scoring. The three
strategies address different aspects of this
problem: ScanTop emphasizes prominent local spans,
PartitionTop preserves evidence from multiple
regions, and WeightedAgg retains broader
response-level coverage. While these complementary
strategies improve detection in many settings,
the current framework does not automatically
determine which strategy is most appropriate for
a particular generation. Developing adaptive
aggregation mechanisms that account for the
characteristics of different reasoning scenarios
is therefore a promising direction.

Moreover, LoRU currently uses fixed window and
selection parameters, selected on a held-out
validation set and shared across experimental
settings. While this design ensures consistent
comparisons, adaptive configurations could better
accommodate variations in reasoning structure,
uncertainty dynamics, and evidence distribution
within individual responses.

Finally, LoRU requires access to token-level
uncertainty signals from the underlying model
or estimator. Although it introduces no additional
model forward passes once these signals are
available, this requirement may limit direct
applicability to black-box systems that do not
expose token-level prediction information.

\section*{Ethical Considerations}

This work aims to improve the trustworthiness of
large language models by identifying potentially
incorrect responses through uncertainty-based
hallucination detection. By improving the assessment
of response reliability, LoRU may help users recognize
unreliable outputs and make more informed decisions
about when additional verification is needed.
Nevertheless, uncertainty-based detection cannot
guarantee factual correctness. Incorrect assessments
may lead users to trust erroneous responses or
unnecessarily reject correct ones. Therefore, LoRU
should serve as a complementary reliability signal
rather than a replacement for external verification
or human judgment, particularly in high-stakes
applications.

\bibliography{custom}

\input{sections/appendix_new}

\end{document}

%% file: sections/introduction.tex
\section{Introduction}
\input{figures/introTeaser}
Large language models (LLMs) can produce convincing responses even when their final answers are incorrect. Uncertainty-based hallucination detection offers a way to assess generation reliability without external knowledge. Prior work has developed richer uncertainty measures through semantic equivalence, relevance-aware weighting, and token-level uncertainty estimation~\citep{farquhar2024detecting,nikitin2024kernel,duan2024shifting,zhang2026tokur}. However, in long, multi-step reasoning, uncertainty signals are distributed across tokens and reasoning steps, raising a central challenge: how should evidence from different parts of a generation be combined to assess final-answer correctness? As illustrated in Figure~\ref{fig:introTeaser}, identical uncertainty scores can have different implications depending on their surrounding uncertainty patterns.

\input{figures/fig_main}

Recent studies have explored uncertainty propagation across generation steps~\citep{zhang2023enhancing,zhao2025uncertainty}, confidence dynamics during reasoning~\citep{yan2026mur,wang2026inference}, and the effects of preceding context or localized confidence on generation reliability~\citep{amit2026evaluating,qiao2026lowest}. These studies highlight the importance of uncertainty beyond isolated predictions. Yet, it remains underexplored how uncertainty scores conditioned on different preceding contexts should be compared, and how their dependence across token positions should inform aggregation.
We investigate two properties relevant to this challenge. First, natural generation and controlled comparisons show that preceding context can substantially shift the uncertainty assigned to the same downstream answer. Second, uncertainty dependence weakens rapidly with token distance across models and benchmarks. These observations motivate two principles: \emph{relativity}, which ranks uncertainty against preceding predictions rather than relying on absolute magnitudes, and \emph{locality}, which restricts this ranking to recent tokens.

Building on these insights, we introduce \textbf{Lo}cal \textbf{R}elative \textbf{U}ncertainty (\textbf{LoRU}), a training-free plug-in framework for uncertainty aggregation. LoRU uses relative ranks to identify uncertainty that is unusually prominent compared with preceding predictions, while sliding windows restrict comparisons to a local range motivated by distance-dependent decay. These ranks guide evidence selection or weighting, while the original uncertainty magnitudes remain responsible for risk scoring.

Local relative ranks identify tokens with elevated uncertainty compared with preceding predictions, but how should this evidence be combined into a response-level score? We explore three aggregation strategies with different evidence coverage. \emph{LoRU-ScanTop} searches the entire response for a contiguous span with the highest average local rank and uses its uncertainty for scoring. \emph{LoRU-PartitionTop} divides the response into partitions and selects the highest-ranked token from each, preserving evidence from different regions. \emph{LoRU-WeightedAgg} retains uncertainty across the response but assigns different weights to sentences based on their local relative ranks. All three can be integrated into existing uncertainty estimators without retraining or additional model forward passes.
Figure~\ref{fig:main_fig} provides an overview of LoRU's local relative ranking and three aggregation strategies.

We evaluate LoRU across three language models, three reasoning benchmarks, and multiple uncertainty estimation methods. Across 65 experimental settings, LoRU improves AUROC in 76.9\% of cases, with an average relative gain of 5.5\%. 
We further demonstrate LoRU's plug-in applicability across four uncertainty estimation approaches with distinct scoring mechanisms, ranging from token-level averaging to step-level spike detection, momentum, and history-based scoring.

Our contributions are threefold: (1) characterizing context-dependent uncertainty shifts and distance-dependent uncertainty relationships; (2) introducing LoRU, a general plug-in framework for local relative uncertainty aggregation; and (3) demonstrating its effectiveness across models, reasoning tasks, and existing uncertainty estimators.

%% file: figures/introTeaser.tex
\begin{figure}[t]
  \centering
  \includegraphics[width=0.9\linewidth]{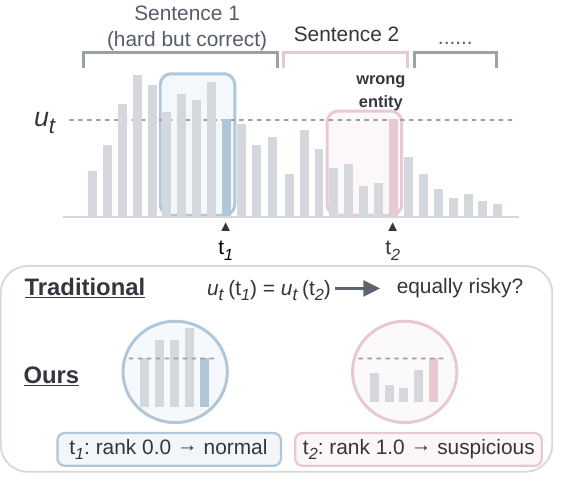}
  \vspace{-0.5em}
\caption{Limitation of absolute uncertainty. Identical uncertainty scores can have different implications depending on their surrounding uncertainty patterns.}
  \label{fig:introTeaser}
  \vspace{-1em}
\end{figure}

%% file: figures/fig_main.tex
\begin{figure*}[t]
  \centering
    \vspace{-2em}
\includegraphics[width=.95\linewidth]{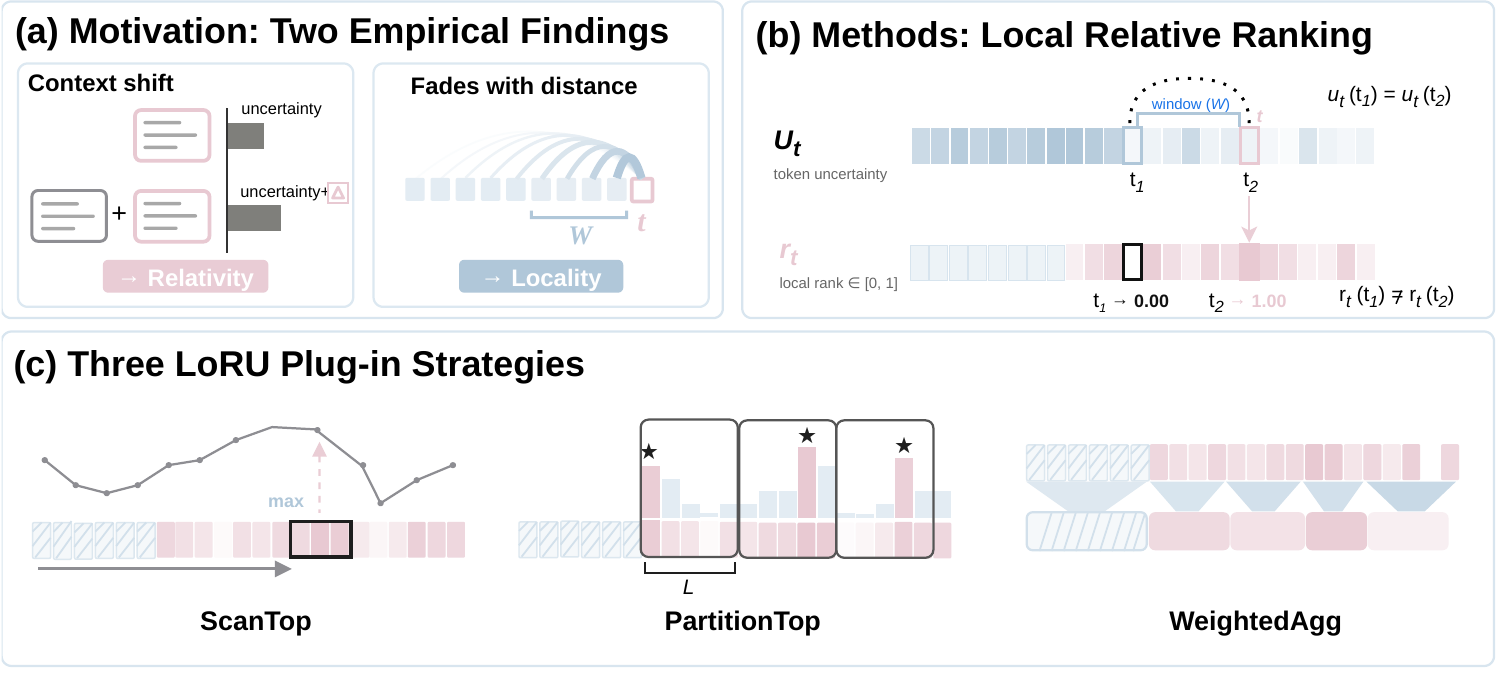}
\vspace{-0.5em}
  \caption{Overview of LoRU. (a) Context shifts uncertainty, and its dependence fades with distance. (b) Each token is ranked against its previous $W$ tokens: $t_1$ and $t_2$ share $u_t$ but rank 0.00 and 1.00. (c) Ranks guide aggregation of the original uncertainties: ScanTop keeps the top-ranked span, PartitionTop the top token per partition, and WeightedAgg weights sentences by mean rank.}
  \vspace{-1em}
  \label{fig:main_fig}
\end{figure*}

%% file: sections/relatedwork.tex
\section{Related Work}
\label{sec:related_work}

\subsection{Uncertainty Estimation and Aggregation}

Uncertainty-based hallucination detection uses signals ranging from token-level uncertainty to response-level semantic consistency. SelfCheckGPT~\cite{manakul2023selfcheckgpt} detects inconsistencies across sampled responses, while Semantic Uncertainty~\cite{kuhn2023semantic} and Semantic Entropy~\cite{farquhar2024detecting} account for semantic equivalence. SAR~\cite{duan2024shifting} weights semantic relevance, Kernel Language Entropy~\cite{nikitin2024kernel} models graded semantic similarities, and GAUSS~\cite{nsgauss} captures discrepancies among factual graphs. Other approaches estimate uncertainty through factual claims~\cite{fadeeva2024fact}, token representations~\cite{vazhentsev2025token}, hidden states~\cite{kossen2024semantic}, supervised uncertainty heads~\cite{shelmanov2025head}, or model perturbations~\cite{zhang2026tokur}.

Beyond uncertainty estimation, response-level aggregation determines how uncertainty evidence contributes to detection. Lowest Span Confidence~\cite{qiao2026lowest} emphasizes low-confidence spans, Rank-Calibration~\cite{huang2024uncertainty} examines uncertainty rankings across responses, and benchmarks~\cite{vashurin2025benchmarking,bakman2025reconsidering} assess estimator effectiveness and robustness. In multi-step reasoning, uncertainty signals span many tokens and reasoning steps, raising the question of how they should be aggregated to assess final-answer correctness.

\subsection{Uncertainty Propagation and Dynamics}

Prior work examines how preceding context affects generation reliability. FOCUS~\cite{zhang2023enhancing} models cascading hallucinations, while semantic-graph methods~\cite{chen2025enhancing} propagate uncertainty through entity and sentence relations. SAUP~\cite{zhao2025uncertainty} aggregates uncertainty across agent trajectories. Lookback Lens~\cite{chuang2024lookback} examines attention to context, SALT~\cite{amit2026evaluating} studies corrupted prefixes and context length, and SHARS~\cite{li2026building} uses segment-wise detection and regeneration to mitigate hallucination accumulation.

Other studies investigate uncertainty dynamics through momentum-based reasoning control~\cite{yan2026mur}, entropy instability~\cite{zhu2026edis}, confidence dynamics and entropy change points~\cite{wang2026inference,xu2026unveiling}, and temporal-logic constraints~\cite{mao2025temporalizing}. Building on these research directions, we characterize context-dependent uncertainty shifts and distance-dependent decay, motivating local relative ranking within sliding windows to guide aggregation. LoRU applies this principle as a training-free plug-in for existing uncertainty estimators.

%% file: sections/understanding.tex
\section{Analyzing Uncertainty Dependence}
\label{sec:uncertainty_propagation}

\subsection{Uncertainty Dependence Across Reasoning Hops}
\label{sec:propagation}

We examine uncertainty dependence across reasoning hops through natural generation and controlled context comparisons.

\paragraph{Natural Generation.}
\textit{Setup.}
We prompt Qwen3.6-27B to generate explicit reasoning hops on DeepSearchQA~\citep{gupta2026deepsearchqa}.
For each hop $h$, we measure in-context uncertainty $U_{\mathrm{ctx}}(h)$ as negative mean token log-probability conditioned on the question and preceding hops, and isolated uncertainty $U_{\mathrm{iso}}(h)$ by scoring the same hop without either.
Figure~\ref{fig:natural_propagation} compares 839 adjacent-hop pairs from 100 responses, grouping previous-hop uncertainty into quartiles and reporting mean current-hop uncertainty with 95\% response-level bootstrap intervals.
Pearson correlations use unbinned pairs.

\input{figures/fig1_natural}

\textit{Findings.}
Previous-hop in-context uncertainty $U_{\mathrm{ctx}}(h-1)$ correlates with current-hop uncertainty $U_{\mathrm{ctx}}(h)$ ($r=0.456$), whereas previous-hop isolated uncertainty $U_{\mathrm{iso}}(h-1)$ shows little correlation with $U_{\mathrm{ctx}}(h)$ ($r=0.048$; Figure~\ref{fig:natural_propagation}).
This contrast suggests that cross-hop uncertainty dependence is not fully captured by scoring individual hops in isolation, although the observational results do not establish causal propagation.
Additional controls are reported in Appendix~\ref{app:natural_controls}.

\paragraph{Controlled Synthetic Intervention.}
\textit{Setup.}
We construct two-hop chains from MuSiQueQA~\citep{trivedi2022musique}, decomposed into $(q_1,a_1)$ and $(q_2,a_2)$.
Keeping both sub-questions fixed, we vary the first-hop answer among Gold, Close, and Far conditions.
GPT-5.1 generates incorrect answers of the same semantic type, excluding candidates overlapping $q_1$ or yielding the gold final answer.
Close and Far candidates are selected by embedding similarity using all-MiniLM-L6-v2.

Given $(q_1,\tilde{a}_1,q_2)$, Llama-3.1-8B generates a second-hop answer $\hat{a}_2$.
We score the same answer under two prompts: an in-context prompt containing the preceding reasoning and an isolated prompt directly asking the resolved question $q_2[\tilde{a}_1]$.
Unlike natural generation, the isolated prompt here retains the resolved second-hop question.
We measure the context-dependent confidence shift as
\begin{equation}
    \Delta = \bar{s}_{\mathrm{ctx}}-\bar{s}_{\mathrm{iso}},
    \label{eq:prefix_effect}
\end{equation}
where $\bar{s}_{\mathrm{ctx}}$ and $\bar{s}_{\mathrm{iso}}$ denote the mean token log-probabilities of the same downstream answer under the two prompts. Positive $\Delta$ indicates higher in-context confidence.
After filtering invalid or context-copied answers, 1,042 chains from 404 questions remain.

\input{figures/fig2_synthetic}

\textit{Findings.}
Figure~\ref{fig:synthetic_propagation} shows context-dependent uncertainty shifts in the same downstream answer, with in-context confidence increasing in 63.7\% of chains (mean $\Delta=+0.30$).
These shifts occur across Gold, Close, and Far conditions, without a reliable difference between Gold and Far.
First-hop isolated uncertainty does not reliably predict $\Delta$ (Appendix~\ref{app:first_hop_prefix}).
These matched comparisons demonstrate context-dependent uncertainty shifts, including when preceding reasoning contains incorrect information, motivating relative uncertainty ranking.

\subsection{Local Decay of Uncertainty Dependence}
\label{sec:decay}

We next examine how uncertainty dependence varies with token distance.

\textit{Setup.}
We analyze token-level full-vocabulary predictive entropy on AIME24--26, MuSiQue, and DeepSearchQA using Llama-3.1-8B-Instruct and Qwen3.5-4B.
For each token distance $d$, we compute correlations between uncertainty values $d$ positions apart.
To distinguish local dependence from global fluctuations, we remove response-specific uncertainty baselines and a shared linear position trend, yielding the controlled correlation $\rho(d)$.
Figure~\ref{fig:uncertainty_decay} compares controlled and raw correlations with 95\% response-level bootstrap intervals, extending to 100-token distances.

\input{figures/fig3_decay}

\textit{Findings.}
Across models and datasets, uncertainty dependence is strongest between nearby tokens and decays rapidly with distance.
Removing global fluctuations makes this local structure more apparent, suggesting that long-range correlations need not indicate sustained dependence.
A hop-level analysis of natural generation shows a consistent decrease in in-context uncertainty correlation with hop distance (Appendix~\ref{app:hop_decay}).
These results motivate restricting uncertainty comparisons to recent predictions rather than the entire generation history.

Together, context-dependent uncertainty shifts and distance-dependent decay motivate LoRU's two complementary principles: relative ranking to account for changing uncertainty levels, and sliding windows to focus comparisons on recent predictions.

%% file: figures/fig1_natural.tex
\begin{figure}[t]
  \centering
  \includegraphics[width=\linewidth]{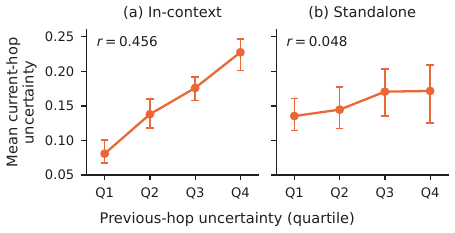}
\caption{Current-hop $U_{\mathrm{ctx}}(h)$ rises with the previous hop's in-context uncertainty (a) but barely with its isolated uncertainty (b). DeepSearchQA, Qwen3.6-27B; 95\% response-level bootstrap intervals.}
  \label{fig:natural_propagation}
  \vspace{-1em}
\end{figure}

%% file: figures/fig2_synthetic.tex
\begin{figure}[t]
  \centering
  \includegraphics[width=\linewidth]{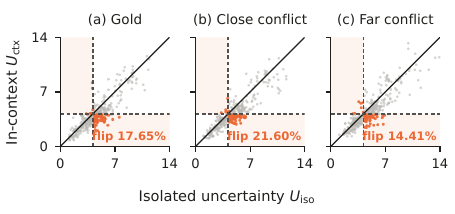}
\caption{Context shifts the same answer's uncertainty: most points lie below $y=x$, and orange points cross a global threshold (dashed), under (a) Gold, (b) Close, and (c) Far first-hop answers.}
\vspace{-1em}
  \label{fig:synthetic_propagation}
\end{figure}

%% file: figures/fig3_decay.tex
\begin{figure*}[t]
  \centering
  \vspace{-1em}
  \includegraphics[width=0.95\linewidth]{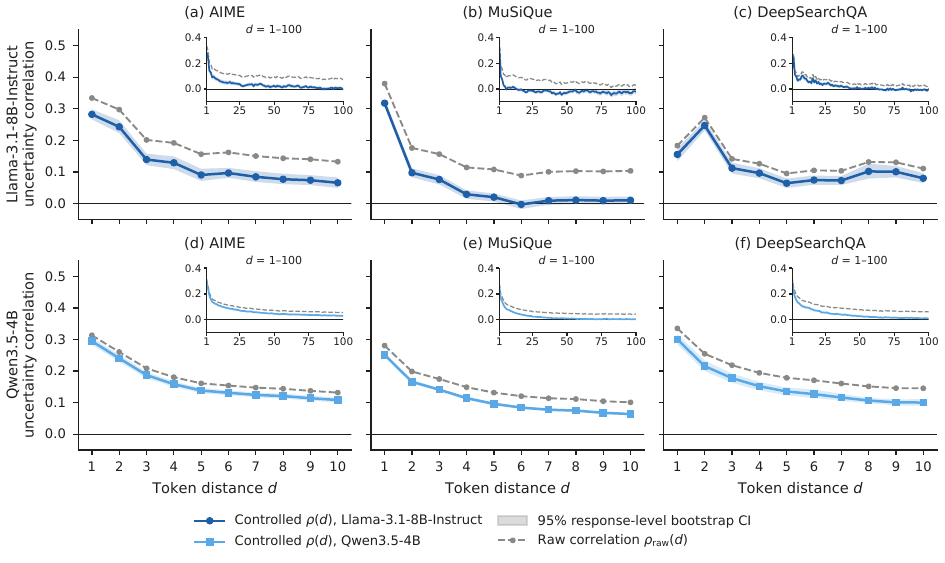}
\vspace{-1em}
\caption{Uncertainty dependence decays within a few tokens, and removing global fluctuations (controlled $\rho(d)$, solid) makes the decay sharper than the raw correlation (dashed). Insets extend the plots ($d\le10$) to $d=100$.}
  \vspace{-1em}\label{fig:uncertainty_decay}
\end{figure*}

%% file: sections/method.tex
\section{LoRU: Local Relative Uncertainty}
\label{sec:relative_uncertainty}

The context-dependent uncertainty shifts and distance-dependent decay observed in Section~\ref{sec:uncertainty_propagation} motivate two complementary principles. \emph{Relativity} ranks uncertainty against preceding values rather than relying on absolute magnitudes alone, while \emph{locality} restricts this ranking to recent tokens. We introduce Local Relative Uncertainty (LoRU), a training-free plug-in framework that uses local relative ranking to guide uncertainty aggregation while preserving the original uncertainty magnitudes for risk scoring. Figure~\ref{fig:main_fig} illustrates these principles and the three rank-guided aggregation strategies.

\subsection{Sliding-Window Locality}
\label{sec:local_reference}

Consider a question $x$, its generated response $y=(y_1,\ldots,y_T)$, and a token-level uncertainty signal $a_t$ used for ranking, where larger values indicate greater uncertainty. Motivated by distance-dependent decay, we define a sliding window of $W$ preceding token positions for each $t>W$:
\begin{equation}
    \mathcal{R}_t=\{t-W,\ldots,t-1\}.
    \label{eq:reference_window}
\end{equation}
The window excludes current and future tokens and advances independently of reasoning-step boundaries. Its size $W$ is specified in Section~\ref{sec:experimental_setup}.

\subsection{Relative Uncertainty Ranking}
\label{sec:relative_ranking}

To account for context-dependent uncertainty shifts, we compare each prediction's uncertainty with its recent history. For $t>W$, we define the normalized local midrank:
\begin{equation}
    r_t=\frac{1}{W}\sum_{i\in\mathcal{R}_t}
    \left[
    \mathbb{I}(a_i<a_t)
    +\frac{1}{2}\mathbb{I}(a_i=a_t)
    \right].
    \label{eq:local_rank}
\end{equation}
Here, $r_t\in[0,1]$ measures the fraction of preceding uncertainty values below $a_t$, with half credit for ties. Higher ranks indicate greater uncertainty relative to recent predictions, providing a local criterion for evidence selection or weighting.

\subsection{Rank-Guided Uncertainty Aggregation}
\label{sec:rank_aggregation}

Let $u_t$ denote the original token-level uncertainty magnitude supplied by an existing estimator. The ranking signal $a_t$ may coincide with $u_t$ or use a different uncertainty measure. LoRU uses local ranks $r_t$ to guide evidence selection or weighting, while retaining $u_t$ for risk scoring.

We instantiate rank-guided aggregation through three complementary strategies: \emph{LoRU-ScanTop} selects uncertainty evidence from a single contiguous region, \emph{LoRU-PartitionTop} selects evidence across multiple regions, and \emph{LoRU-WeightedAgg} adjusts evidence contributions through soft weighting. Below, we present their canonical forms for token-mean estimators.

\paragraph{LoRU-ScanTop: Top-Span Selection.}
ScanTop searches overlapping spans of length $B$ and selects the one with the highest average local rank. For $I_b=\{b,\ldots,b+B-1\}$, where $W<b\le T-B+1$,
\begin{equation}
    q_b=\frac{1}{B}\sum_{t\in I_b}r_t,
    \qquad
    b^*=\min\arg\max_b q_b.
    \label{eq:scantop}
\end{equation}
The selected span is scored using the original uncertainty magnitudes:
\begin{equation}
    S_{\mathrm{ScanTop}}(x,y)=
    \frac{1}{B}\sum_{t\in I_{b^*}}u_t.
    \label{eq:scantop_score}
\end{equation}
Thus, ScanTop searches across the response but concentrates scoring on one contiguous region.

\paragraph{LoRU-PartitionTop: Partition-Wise Selection.}
PartitionTop divides eligible positions into consecutive, non-overlapping partitions $C_1,\ldots,C_J$ of length $L$. It selects the highest-ranked token in each partition:
\begin{equation}
    t_j^*=\min\arg\max_{t\in C_j}r_t,
    \qquad j=1,\ldots,J.
    \label{eq:partitiontop}
\end{equation}
The response score averages the original uncertainty magnitudes of selected tokens:
\begin{equation}
    S_{\mathrm{PartitionTop}}(x,y)=
    \frac{1}{J}\sum_{j=1}^{J}u_{t_j^*}.
    \label{eq:partitiontop_score}
\end{equation}
Unlike ScanTop, PartitionTop preserves representative uncertainty evidence across different regions of the response.

\paragraph{LoRU-WeightedAgg: Sentence-Level Weighting.}
Unlike ScanTop and PartitionTop, which use token-wise local relative ranks for evidence selection, WeightedAgg uses sentence-anchored local relative ranks to softly weight uncertainty evidence across the response.

Partition the response into sentences $S_1,\ldots,S_J$, with sentence $S_j$ starting at token $\tau_j$. For $\tau_j>W$, all tokens $t\in S_j$ are compared against the same preceding window $\mathcal{R}_{\tau_j}$:
\begin{equation}
    r_{t\mid j}=\frac{1}{W}
    \sum_{i\in\mathcal{R}_{\tau_j}}
    \left[
    \mathbb{I}(a_i<a_t)
    +\frac{1}{2}\mathbb{I}(a_i=a_t)
    \right], t\in S_j.
    \label{eq:sentence_anchor_rank}
\end{equation}
Each sentence receives weight
$w_j=|S_j|^{-1}\sum_{t\in S_j}r_{t\mid j}$.
The resulting response score is
\begin{equation}
    S_{\mathrm{WeightedAgg}}(x,y)
    =\frac{\sum_{j=1}^{J}w_j\sum_{t\in S_j}u_t}
    {\sum_{j=1}^{J}w_j|S_j|}.
    \label{eq:weightedagg_score}
\end{equation}
WeightedAgg therefore adjusts evidence contributions without excluding unselected positions.

\paragraph{Plug-in Integration and Cost.}
LoRU modifies uncertainty aggregation without changing the underlying uncertainty estimator. While the above equations define token-mean aggregation, LoRU can also modify corresponding selection or weighting operations in step-level estimators while retaining their native scoring structures. The operator-matched implementations are described in Section~\ref{sec:experimental_setup}, with boundary conditions and implementation details in Appendix~\ref{app:loru_implementation}. Given precomputed uncertainty signals, local rank computation requires $O(TW)$ comparisons and no additional model forward passes.

%% file: sections/experiments.tex
\section{LoRU Experiments}
\label{sec:experiments}

We evaluate LoRU as a plug-in for response-level hallucination detection, comparing each variant with its native estimator on identical generated responses and uncertainty measurements. We examine detection effectiveness, plug-in applicability, and aggregation strategies.

\subsection{Experimental Setup}
\label{sec:experimental_setup}

\paragraph{Task and Datasets.}
Given a question $x$ and generated response $y$, the task is to distinguish incorrect from correct final answers using a response-level risk score $S(x,y)$, where higher scores indicate greater risk. We evaluate three reasoning benchmarks: AIME24--26 (90 mathematics problems), MuSiQue-4hop~\citep{trivedi2022musique} (458 four-hop factual questions), and DeepSearchQA~\citep{gupta2026deepsearchqa} (316 single-answer information-seeking questions). MuSiQue-4hop and DeepSearchQA are evaluated without external retrieval.

\paragraph{Models and Baselines.}
We use Llama-3.1-8B-Instruct~\citep{grattafiori2024llama}, Qwen3.5-4B, and Qwen3.8-27B. Each model generates one reasoning response per question. Qwen uses temperature $1.0$, top-$p$ $0.95$, and top-$k$ $20$, while Llama uses greedy decoding.

We evaluate five response-level scoring configurations from four uncertainty estimation approaches. \emph{TokUR-EU-Mean}~\citep{zhang2026tokur} averages perturbation-based token epistemic uncertainty, while \emph{LL} averages token negative log-likelihood. \emph{MUR-Spike} and \emph{MUR-Momentum}~\citep{yan2026mur} use the maximum step-level uncertainty increase and final bias-corrected momentum, respectively. \emph{Avg. Uncertainty} compares step uncertainty with its preceding arithmetic average and applies mean-based response scoring. These baselines cover token-level and step-level uncertainty aggregation.

\paragraph{LoRU Plug-in Integration.}
Following Section~\ref{sec:rank_aggregation}, LoRU computes local relative ranks from the ranking signal $a_t$ while retaining the original uncertainty magnitudes $u_t$ for risk scoring. TokUR uses token epistemic uncertainty as $u_t$; LL, MUR, and Avg. Uncertainty use token negative log-likelihood. ScanTop and PartitionTop use unperturbed predictive entropy as $a_t$, while WeightedAgg uses the corresponding estimator's uncertainty signal.

For token-mean estimators (TokUR and LL), all three strategies directly modify response-level aggregation. For step-level estimators (MUR and Avg. Uncertainty), operator-matched integration preserves their step-level scoring structures: PartitionTop modifies within-step token aggregation, ScanTop modifies prominent-step selection in MUR-Spike, and WeightedAgg modifies cross-step weighting in MUR-Momentum and Avg. Uncertainty. PartitionTop is evaluated with all step-level scores.

\input{tables/exp_mean}

\paragraph{Implementation and Evaluation.}
LoRU hyperparameters are selected on a held-out validation set and fixed across models, datasets, and estimators: window size $W=20$, ScanTop span length $B=5$, and PartitionTop partition length $L=20$. Further details are provided in Appendix~\ref{app:experiment_details}.
Paired comparisons use $2{,}000$ question-level bootstrap resamples, with confidence intervals reported for the controlled comparisons and ablations (Section~\ref{sec:strategy_analysis}; Appendices~\ref{app:aggregation_analysis} and~\ref{app:ablation}).

We report AUROC, AUPRC, and prediction rejection ratio (PRR), all scaled by $100$. AUROC treats incorrect answers as positive and ranks them by risk; AUPRC treats correct answers as positive and ranks them by negative risk. PRR evaluates accuracy as high-risk responses are rejected, normalized between random ($0$) and oracle ($100$) rejection. All metrics are threshold-free. Paired comparisons use $2{,}000$ question-level bootstrap resamples, with confidence intervals reported in the Appendix.

\subsection{Overall Effectiveness and Computational Cost}
\label{sec:overall_effectiveness}

\paragraph{Overall Detection Effectiveness.}
Table~\ref{tab:main_results} compares LoRU with native estimators across models, datasets, and scoring configurations. Across 65 experimental settings, the best-performing evaluated LoRU strategy improves AUROC in 76.9\% of cases, with an average relative gain of 5.5\%. This per-setting best-variant comparison is descriptive rather than the performance of a fixed strategy.

The strongest cross-estimator consistency appears on MuSiQue-4hop, where PartitionTop improves AUROC in 14 of 15 model--estimator comparisons. Other strategies provide complementary improvements on mathematical reasoning and information-seeking tasks, although effectiveness varies across configurations. These results show that local relative ranking can improve hallucination detection by modifying uncertainty aggregation across different estimators.

\paragraph{Computational Cost.}
LoRU requires no retraining or additional model forward passes beyond those of the underlying estimator. Given the required uncertainty signals, computing local relative ranks takes $O(TW)$ comparisons, followed by evidence selection or weighting. Thus, LoRU adds lightweight aggregation computation without changing model inference.

\input{tables/loru_commitment}

\subsection{Analysis of Aggregation Strategies}
\label{sec:strategy_analysis}

\paragraph{Local Relative Ranks at Factual Commitments.}
We examine the first generated mentions of matched MuSiQue hop sub-answers, excluding copies from the question. Table~\ref{tab:loru_commitment} shows their elevated predictive entropy relative to the response median and their local relative ranks. By contrast, mathematical reasoning contains many low-entropy numerical and symbolic tokens. These differences illustrate heterogeneous uncertainty distributions across reasoning tasks, motivating aggregation strategies that preserve evidence from different response regions.

\paragraph{Evidence Coverage Beyond Selection Budget.}
ScanTop retains one contiguous span for scoring, whereas PartitionTop selects evidence across multiple partitions. TokUR-EU controls (Table~\ref{tab:loru_selection_controls}) distinguish selection budget from evidence coverage. Increasing the number of retained tokens improves AUROC, while partition-wise selection retains a $+2.0\%$ advantage over global selection after matching both the selection budget and unit. This suggests that distributing evidence across the response can provide benefits beyond retaining more tokens.

\input{tables/controlled_comp}

\paragraph{Ranking--Scoring Alignment.}
LoRU uses local relative ranks for evidence selection or weighting while retaining the original uncertainty magnitudes for risk scoring. The relationship between these signals may affect aggregation effectiveness.

On MuSiQue TokUR-EU, ScanTop improves AUROC on both Qwen models but decreases it on Llama. This difference coincides with positive within-response Spearman correlations between predictive entropy and epistemic uncertainty on Qwen ($0.92$--$0.97$), compared with a negative correlation on Llama ($-0.14$). This suggests that ranking--scoring alignment matters for concentrated evidence selection, without establishing causality.

PartitionTop reduces reliance on a single region, while WeightedAgg retains response-wide evidence through soft weighting. Together, these analyses highlight evidence coverage and ranking--scoring alignment as two considerations for rank-guided aggregation, with no universally superior strategy. Additional analyses are provided in Appendix~\ref{app:aggregation_analysis}.

\paragraph{Ablation Study.}
We conduct a $2\times2$ ablation of LoRU-ScanTop on TokUR-EU across nine model--dataset settings. Combining a local window with relative ranking yields higher mean AUROC than either windowed z-scoring or full-prefix ranking, with a positive interaction of $+4.7$ points (95\% CI: $[+0.6,+8.8]$). These results support their complementarity in this setting. Full ablation results are provided in Appendix~\ref{app:ablation}.

%% file: tables/exp_mean.tex
\begin{table*}[!t]
\vspace{-2em}
\centering
\definecolor{gain}{HTML}{2F6FD0}
\resizebox{\textwidth}{!}{%
\begin{tabular}{l ccc ccc ccc ccc ccc}
\toprule
\multirow{2}{*}{\textbf{Method}}
  & \multicolumn{3}{c}{\textbf{MuSiQue-4hop}}
  & \multicolumn{3}{c}{\textbf{AIME26}}
  & \multicolumn{3}{c}{\textbf{AIME25}}
  & \multicolumn{3}{c}{\textbf{AIME24}}
  & \multicolumn{3}{c}{\textbf{DeepSearchQA}} \\
\cmidrule(lr){2-4}\cmidrule(lr){5-7}\cmidrule(lr){8-10}\cmidrule(lr){11-13}\cmidrule(lr){14-16}
  & AUROC & AUPRC & PRR
  & AUROC & AUPRC & PRR
  & AUROC & AUPRC & PRR
  & AUROC & AUPRC & PRR
  & AUROC & AUPRC & PRR \\
\midrule
\multicolumn{16}{c}{\texttt{Llama-3.1-8B-Instruct}} \\
\midrule
\texttt{TokUR} (EU-Mean) & 50.79 & 19.60 & -3.18 & -- & -- & -- & -- & -- & -- & 68.89 & 24.42 & 23.71 & 59.70 & \textbf{12.77} & \textbf{13.48} \\
\texttt{LoRU-ScanTop} & 47.71 & 18.68 & -5.45 & -- & -- & -- & -- & -- & -- & \cellcolor{gain!40}\textbf{82.22} & \cellcolor{gain!40}\textbf{37.50} & \cellcolor{gain!33}\textbf{45.48} & \cellcolor{gain!25}\underline{61.65} & 12.29 & 12.05 \\
\texttt{LoRU-PartitionTop} & \cellcolor{gain!28}\textbf{52.62} & \cellcolor{gain!24}\textbf{20.84} & \cellcolor{gain!26}\textbf{1.21} & -- & -- & -- & -- & -- & -- & \cellcolor{gain!25}\underline{75.56} & \cellcolor{gain!19}27.87 & \cellcolor{gain!19}31.55 & \cellcolor{gain!34}\textbf{63.02} & \underline{12.45} & \underline{12.82} \\
\texttt{LoRU-WeightedAgg} & \cellcolor{gain!9}\underline{50.85} & \cellcolor{gain!8}\underline{19.70} & \cellcolor{gain!9}\underline{-2.93} & -- & -- & -- & -- & -- & -- & \cellcolor{gain!14}73.33 & \cellcolor{gain!27}\underline{29.42} & \cellcolor{gain!20}\underline{32.74} & \cellcolor{gain!8}59.76 & \underline{12.45} & 12.72 \\
\midrule
$\texttt{LL}$ & 51.89 & 21.62 & 3.61 & -- & -- & -- & -- & -- & -- & 66.67 & 22.82 & 18.83 & \textbf{60.85} & \textbf{11.63} & \textbf{9.53} \\
\texttt{LoRU-ScanTop} & \cellcolor{gain!34}\textbf{56.30} & \cellcolor{gain!27}\textbf{22.93} & \cellcolor{gain!23}\textbf{7.53} & -- & -- & -- & -- & -- & -- & \cellcolor{gain!23}\underline{73.33} & \cellcolor{gain!17}\underline{25.60} & \cellcolor{gain!17}\underline{26.43} & 55.91 & \underline{10.94} & 6.15 \\
\texttt{LoRU-PartitionTop} & \cellcolor{gain!20}\underline{52.79} & \cellcolor{gain!15}22.15 & \cellcolor{gain!8}3.75 & -- & -- & -- & -- & -- & -- & 66.67 & 20.60 & 4.70 & 52.64 & 9.28 & 0.82 \\
\texttt{LoRU-WeightedAgg} & \cellcolor{gain!19}52.74 & \cellcolor{gain!21}\underline{22.58} & \cellcolor{gain!18}\underline{6.35} & -- & -- & -- & -- & -- & -- & \cellcolor{gain!31}\textbf{75.56} & \cellcolor{gain!34}\textbf{30.10} & \cellcolor{gain!31}\textbf{34.56} & \underline{59.24} & 10.79 & \underline{6.75} \\
\midrule
$\texttt{Avg. uncertainty (Mean)}$ & 51.95 & 22.29 & 4.04 & -- & -- & -- & -- & -- & -- & \underline{73.33} & 24.17 & \underline{19.74} & \underline{60.50} & \textbf{11.27} & \textbf{8.44} \\
$\texttt{LoRU-PartitionTop}$ & \cellcolor{gain!31}\textbf{55.13} & \cellcolor{gain!20}\underline{23.13} & \cellcolor{gain!25}\textbf{8.08} & -- & -- & -- & -- & -- & -- & \cellcolor{gain!11}\textbf{75.56} & \cellcolor{gain!29}\textbf{30.10} & \cellcolor{gain!28}\textbf{34.56} & \cellcolor{gain!23}\textbf{61.98} & \underline{11.24} & \underline{7.32} \\
$\texttt{LoRU-WeightedAgg}$ & \cellcolor{gain!17}\underline{52.61} & \cellcolor{gain!28}\textbf{23.76} & \cellcolor{gain!22}\underline{7.80} & -- & -- & -- & -- & -- & -- & 71.11 & \cellcolor{gain!10}\underline{24.27} & 19.52 & 59.22 & 10.69 & 6.79 \\
\midrule
\texttt{MUR (spike)} & \underline{54.21} & \textbf{26.26} & \textbf{12.89} & -- & -- & -- & -- & -- & -- & 62.22 & 19.05 & -2.18 & \textbf{61.85} & \textbf{12.59} & \textbf{11.85} \\
\texttt{LoRU-ScanTop} & 49.62 & 18.93 & -5.72 & -- & -- & -- & -- & -- & -- & \cellcolor{gain!25}\textbf{68.89} & \cellcolor{gain!23}\textbf{22.88} & \cellcolor{gain!29}\textbf{12.83} & 55.39 & 9.46 & 0.97 \\
\texttt{LoRU-PartitionTop} & \cellcolor{gain!26}\textbf{55.94} & \underline{22.90} & \underline{7.32} & -- & -- & -- & -- & -- & -- & \cellcolor{gain!17}\underline{66.67} & \cellcolor{gain!16}\underline{20.97} & \cellcolor{gain!24}\underline{12.16} & \underline{58.54} & \underline{10.92} & \underline{6.22} \\
\midrule
\texttt{MUR (momentum)} & 51.40 & 22.29 & 4.09 & -- & -- & -- & -- & -- & -- & \underline{80.00} & \underline{30.18} & \underline{36.29} & \underline{61.92} & \textbf{11.66} & \textbf{9.44} \\
\texttt{LoRU-PartitionTop} & \cellcolor{gain!33}\textbf{55.38} & \cellcolor{gain!23}\underline{23.36} & \cellcolor{gain!27}\textbf{8.54} & -- & -- & -- & -- & -- & -- & \cellcolor{gain!14}\textbf{84.44} & \cellcolor{gain!25}\textbf{34.17} & \cellcolor{gain!15}\textbf{42.50} & \cellcolor{gain!20}\textbf{63.20} & \underline{11.65} & \underline{8.80} \\
\texttt{LoRU-WeightedAgg} & \cellcolor{gain!18}\underline{52.24} & \cellcolor{gain!25}\textbf{23.59} & \cellcolor{gain!20}\underline{7.10} & -- & -- & -- & -- & -- & -- & \cellcolor{gain!14}\textbf{84.44} & \cellcolor{gain!25}\textbf{34.17} & \cellcolor{gain!15}\textbf{42.50} & 59.61 & 10.83 & 7.03 \\
\midrule
\multicolumn{16}{c}{\texttt{Qwen-3.5-4B}} \\
\midrule
\texttt{TokUR} (EU-Mean) & 42.36 & 5.50 & -8.40 & \underline{81.70} & \underline{70.24} & \underline{42.09} & 60.99 & \underline{57.11} & \underline{19.17} & \underline{87.14} & 96.18 & 85.36 & \underline{63.79} & \textbf{15.97} & \textbf{17.76} \\
\texttt{LoRU-ScanTop} & \cellcolor{gain!39}\textbf{50.52} & \cellcolor{gain!30}\underline{6.98} & \cellcolor{gain!39}\textbf{0.51} & 59.82 & 66.73 & 30.04 & 36.26 & 38.33 & -33.94 & 35.71 & 72.51 & -13.38 & 53.65 & 12.40 & 6.83 \\
\texttt{LoRU-PartitionTop} & \cellcolor{gain!23}\underline{43.69} & \cellcolor{gain!31}\textbf{6.99} & \cellcolor{gain!36}\underline{-0.49} & 79.91 & 68.11 & 32.42 & \cellcolor{gain!32}\textbf{64.84} & 54.20 & 15.54 & \cellcolor{gain!8}\textbf{87.86} & \cellcolor{gain!12}\underline{96.40} & \cellcolor{gain!10}\underline{86.20} & 61.10 & 14.39 & 13.22 \\
\texttt{LoRU-WeightedAgg} & \cellcolor{gain!10}42.50 & \cellcolor{gain!9}5.61 & \cellcolor{gain!13}-7.52 & \cellcolor{gain!26}\textbf{84.38} & \cellcolor{gain!31}\textbf{74.07} & \cellcolor{gain!26}\textbf{50.72} & \cellcolor{gain!8}\underline{61.54} & \cellcolor{gain!26}\textbf{59.20} & \cellcolor{gain!22}\textbf{24.34} & \cellcolor{gain!8}\textbf{87.86} & \cellcolor{gain!14}\textbf{96.44} & \cellcolor{gain!13}\textbf{86.30} & \cellcolor{gain!11}\textbf{63.92} & \underline{15.93} & \underline{17.61} \\
\midrule
$\texttt{LL}$ & 48.08 & 6.20 & -4.01 & \underline{80.80} & \underline{76.37} & \underline{55.79} & \underline{62.09} & \underline{56.48} & \underline{15.74} & \underline{90.71} & \underline{96.94} & \underline{88.73} & \underline{63.20} & \underline{15.18} & 15.40 \\
\texttt{LoRU-ScanTop} & \cellcolor{gain!23}\underline{49.41} & \cellcolor{gain!21}\textbf{7.16} & \cellcolor{gain!28}\textbf{0.73} & 45.54 & 46.85 & -26.77 & 31.32 & 36.86 & -41.47 & 35.71 & 71.27 & -17.97 & 55.90 & 11.34 & 2.51 \\
\texttt{LoRU-PartitionTop} & \cellcolor{gain!35}\textbf{52.76} & \cellcolor{gain!17}\underline{6.86} & \cellcolor{gain!21}\underline{-0.76} & 78.57 & 73.37 & 49.32 & \cellcolor{gain!16}\textbf{63.19} & 52.04 & 10.21 & \cellcolor{gain!8}\textbf{91.43} & \cellcolor{gain!12}\textbf{97.16} & \cellcolor{gain!12}\textbf{89.59} & \cellcolor{gain!31}\textbf{66.03} & \cellcolor{gain!8}\textbf{15.32} & \cellcolor{gain!11}\textbf{15.68} \\
\texttt{LoRU-WeightedAgg} & \cellcolor{gain!14}48.63 & 6.05 & -5.57 & \cellcolor{gain!12}\textbf{81.70} & \cellcolor{gain!19}\textbf{77.92} & \cellcolor{gain!17}\textbf{58.85} & 60.99 & \cellcolor{gain!8}\textbf{57.29} & \cellcolor{gain!17}\textbf{19.24} & 87.86 & 96.05 & 85.23 & 62.33 & 15.10 & \cellcolor{gain!8}\underline{15.59} \\
\midrule
$\texttt{Avg. uncertainty (Mean)}$ & \underline{49.36} & \textbf{7.00} & \underline{-1.33} & \underline{75.00} & 66.75 & 34.47 & \textbf{68.13} & \underline{57.18} & \underline{22.01} & \underline{80.71} & \underline{90.65} & \underline{65.79} & \underline{61.73} & \underline{14.55} & \underline{13.25} \\
$\texttt{LoRU-PartitionTop}$ & \cellcolor{gain!30}\textbf{52.04} & \underline{6.95} & \cellcolor{gain!16}\textbf{0.69} & 73.66 & \cellcolor{gain!26}\underline{69.60} & \cellcolor{gain!22}\underline{40.35} & \underline{66.48} & \cellcolor{gain!17}\textbf{58.34} & \cellcolor{gain!13}\textbf{25.08} & 72.86 & 85.43 & 45.25 & \cellcolor{gain!28}\textbf{63.74} & \cellcolor{gain!22}\textbf{14.85} & \cellcolor{gain!30}\textbf{14.70} \\
$\texttt{LoRU-WeightedAgg}$ & 47.40 & 6.11 & -4.30 & \cellcolor{gain!33}\textbf{81.70} & \cellcolor{gain!35}\textbf{76.95} & \cellcolor{gain!35}\textbf{57.02} & 62.09 & 56.50 & 15.83 & \cellcolor{gain!37}\textbf{90.71} & \cellcolor{gain!31}\textbf{96.77} & \cellcolor{gain!35}\textbf{88.28} & 59.13 & 14.08 & 11.63 \\
\midrule
\texttt{MUR (spike)} & \underline{46.44} & \underline{6.08} & \underline{-5.33} & \underline{57.59} & \underline{54.35} & \underline{-1.70} & \textbf{63.74} & \underline{54.63} & \underline{16.78} & \underline{63.57} & \underline{83.89} & \underline{38.67} & 50.71 & 10.52 & -0.50 \\
\texttt{LoRU-ScanTop} & 43.69 & 5.72 & -7.76 & \cellcolor{gain!40}\textbf{80.36} & \cellcolor{gain!40}\textbf{84.80} & \cellcolor{gain!40}\textbf{70.53} & \underline{58.79} & \cellcolor{gain!31}\textbf{57.14} & \cellcolor{gain!8}\textbf{18.72} & \cellcolor{gain!20}\textbf{70.00} & \cellcolor{gain!21}\textbf{87.53} & \cellcolor{gain!26}\textbf{53.12} & \cellcolor{gain!40}\textbf{58.12} & \cellcolor{gain!36}\textbf{12.46} & \cellcolor{gain!40}\textbf{7.20} \\
\texttt{LoRU-PartitionTop} & \cellcolor{gain!38}\textbf{52.95} & \cellcolor{gain!18}\textbf{6.89} & \cellcolor{gain!30}\textbf{-0.20} & 49.55 & 52.26 & -3.63 & 58.24 & 53.52 & 13.51 & 57.14 & 82.04 & 29.87 & \cellcolor{gain!37}\underline{55.27} & \cellcolor{gain!29}\underline{11.72} & \cellcolor{gain!37}\underline{4.09} \\
\midrule
\texttt{MUR (momentum)} & \underline{48.98} & \underline{6.86} & \underline{-2.10} & \underline{78.57} & 70.40 & 42.85 & \textbf{68.13} & \textbf{55.59} & \textbf{18.29} & \underline{78.57} & \underline{87.01} & \underline{52.13} & \underline{60.69} & 13.17 & \underline{9.83} \\
\texttt{LoRU-PartitionTop} & \cellcolor{gain!22}\textbf{50.25} & \cellcolor{gain!14}\textbf{7.15} & \cellcolor{gain!17}\textbf{0.10} & 75.00 & \cellcolor{gain!10}\underline{71.01} & \cellcolor{gain!10}\underline{43.96} & \underline{66.48} & \underline{55.53} & \underline{14.17} & 73.57 & 84.32 & 41.04 & \cellcolor{gain!17}\textbf{61.94} & \cellcolor{gain!26}\textbf{14.09} & \cellcolor{gain!34}\textbf{11.71} \\
\texttt{LoRU-WeightedAgg} & 46.59 & 6.60 & -3.82 & \cellcolor{gain!19}\textbf{80.36} & \cellcolor{gain!33}\textbf{75.70} & \cellcolor{gain!31}\textbf{54.28} & 62.64 & 51.29 & 6.25 & \cellcolor{gain!28}\textbf{86.43} & \cellcolor{gain!36}\textbf{95.41} & \cellcolor{gain!36}\textbf{82.98} & 58.38 & \cellcolor{gain!15}\underline{13.42} & \underline{9.83} \\
\midrule
\multicolumn{16}{c}{\texttt{Qwen-3.8-27B}} \\
\midrule
\texttt{TokUR} (EU-Mean) & 37.65 & 10.18 & -15.18 & \underline{75.32} & 90.92 & 63.59 & \underline{59.21} & 69.67 & -7.25 & \underline{69.57} & 82.39 & 11.70 & \textbf{55.99} & 20.40 & 11.33 \\
\texttt{LoRU-ScanTop} & \cellcolor{gain!40}\textbf{54.64} & \cellcolor{gain!40}\textbf{13.94} & \cellcolor{gain!40}\textbf{0.73} & 60.39 & 82.89 & 29.28 & 55.26 & \cellcolor{gain!35}\textbf{72.22} & \cellcolor{gain!35}\textbf{2.55} & 63.77 & \cellcolor{gain!32}\textbf{88.79} & \cellcolor{gain!38}\textbf{45.88} & 50.76 & \cellcolor{gain!40}\textbf{22.36} & \cellcolor{gain!27}\textbf{12.72} \\
\texttt{LoRU-PartitionTop} & \cellcolor{gain!29}\underline{40.13} & \cellcolor{gain!12}\underline{10.43} & \cellcolor{gain!12}-14.69 & \cellcolor{gain!15}\textbf{76.62} & \cellcolor{gain!17}\textbf{92.46} & \cellcolor{gain!19}\textbf{69.33} & 58.55 & 69.02 & -15.67 & 61.59 & 79.55 & -2.82 & 52.72 & \cellcolor{gain!33}\underline{21.84} & \cellcolor{gain!14}\underline{11.64} \\
\texttt{LoRU-WeightedAgg} & \cellcolor{gain!15}38.23 & \cellcolor{gain!11}10.35 & \cellcolor{gain!14}\underline{-13.98} & \cellcolor{gain!15}\textbf{76.62} & \cellcolor{gain!15}\underline{91.67} & \cellcolor{gain!15}\underline{66.62} & \cellcolor{gain!24}\textbf{60.53} & \cellcolor{gain!13}\underline{70.78} & \cellcolor{gain!26}\underline{-1.22} & \cellcolor{gain!8}\textbf{70.29} & \cellcolor{gain!8}\underline{82.45} & \cellcolor{gain!8}\underline{12.09} & \underline{54.41} & 19.42 & 8.83 \\
\midrule
$\texttt{LL}$ & 49.09 & \underline{14.68} & \underline{3.17} & \underline{70.78} & 88.96 & 55.38 & \textbf{57.24} & \underline{68.99} & \underline{-9.81} & \textbf{75.36} & \underline{84.03} & \underline{19.71} & \textbf{59.53} & \underline{24.12} & \underline{16.58} \\
\texttt{LoRU-ScanTop} & \cellcolor{gain!13}\textbf{49.59} & 12.95 & -2.44 & 54.55 & 80.16 & 16.30 & 48.68 & 64.61 & -30.95 & 57.97 & \cellcolor{gain!16}\textbf{85.95} & \cellcolor{gain!22}\textbf{31.29} & 50.73 & 19.78 & 7.88 \\
\texttt{LoRU-PartitionTop} & 47.73 & 14.20 & 1.49 & \cellcolor{gain!8}\textbf{71.43} & \cellcolor{gain!13}\textbf{89.63} & \cellcolor{gain!13}\textbf{58.03} & \underline{56.58} & 67.90 & -20.04 & 72.46 & 83.29 & 16.07 & \underline{59.39} & \cellcolor{gain!19}\textbf{24.40} & \cellcolor{gain!21}\textbf{17.07} \\
\texttt{LoRU-WeightedAgg} & \cellcolor{gain!8}\underline{49.12} & \cellcolor{gain!12}\textbf{14.93} & \cellcolor{gain!11}\textbf{3.51} & \cellcolor{gain!8}\textbf{71.43} & \cellcolor{gain!8}\underline{89.20} & \cellcolor{gain!8}\underline{56.43} & \textbf{57.24} & \cellcolor{gain!22}\textbf{70.25} & \cellcolor{gain!31}\textbf{-2.60} & \underline{73.19} & 83.52 & 17.16 & 57.84 & 22.77 & 14.45 \\
\midrule
$\texttt{Avg. uncertainty (Mean)}$ & \underline{48.11} & 12.14 & -6.65 & \underline{77.27} & \underline{90.73} & \underline{63.10} & \textbf{57.89} & \textbf{72.40} & \textbf{5.36} & \textbf{86.96} & \textbf{95.20} & \textbf{78.07} & \textbf{58.95} & \textbf{25.69} & \underline{18.00} \\
$\texttt{LoRU-PartitionTop}$ & \cellcolor{gain!24}\textbf{49.46} & \cellcolor{gain!39}\textbf{14.68} & \cellcolor{gain!34}\underline{0.61} & \cellcolor{gain!8}\textbf{77.92} & \cellcolor{gain!22}\textbf{92.50} & \cellcolor{gain!24}\textbf{69.88} & 52.63 & 66.29 & -26.49 & 58.70 & 79.82 & -1.74 & \underline{58.77} & \underline{24.65} & \cellcolor{gain!18}\textbf{18.40} \\
$\texttt{LoRU-WeightedAgg}$ & 47.93 & \cellcolor{gain!33}\underline{14.10} & \cellcolor{gain!35}\textbf{0.88} & 69.48 & 88.38 & 52.94 & \underline{53.95} & \underline{69.85} & \underline{-4.18} & \underline{75.36} & \underline{84.03} & \underline{19.71} & 58.16 & 23.11 & 14.01 \\
\midrule
\texttt{MUR (spike)} & \underline{47.01} & \underline{13.69} & \underline{-1.91} & 55.52 & 76.67 & -1.49 & \underline{42.11} & \underline{63.60} & \underline{-31.70} & \underline{55.80} & \underline{79.51} & \underline{-8.27} & \underline{54.74} & \underline{20.00} & \underline{7.96} \\
\texttt{LoRU-ScanTop} & 46.48 & 12.69 & -3.47 & \cellcolor{gain!36}\textbf{68.18} & \cellcolor{gain!38}\textbf{89.82} & \cellcolor{gain!38}\textbf{57.77} & \cellcolor{gain!40}\textbf{61.18} & \cellcolor{gain!40}\textbf{80.74} & \cellcolor{gain!40}\textbf{35.94} & \cellcolor{gain!34}\textbf{65.22} & \cellcolor{gain!38}\textbf{88.17} & \cellcolor{gain!40}\textbf{43.40} & 52.37 & 17.79 & 3.91 \\
\texttt{LoRU-PartitionTop} & \cellcolor{gain!36}\textbf{52.99} & \cellcolor{gain!34}\textbf{15.67} & \cellcolor{gain!32}\textbf{4.93} & \cellcolor{gain!29}\underline{62.01} & \cellcolor{gain!29}\underline{79.68} & \cellcolor{gain!33}\underline{12.40} & 34.87 & 60.20 & -52.80 & 54.35 & 79.03 & -10.80 & \cellcolor{gain!14}\textbf{55.95} & \cellcolor{gain!12}\textbf{20.16} & \cellcolor{gain!24}\textbf{8.58} \\
\midrule
\texttt{MUR (momentum)} & 48.14 & 12.09 & -6.97 & \underline{77.27} & \underline{90.67} & \underline{62.92} & \textbf{60.53} & \textbf{73.77} & \textbf{10.25} & \textbf{84.06} & \textbf{94.54} & \textbf{74.86} & \textbf{59.80} & \textbf{26.15} & \textbf{19.20} \\
\texttt{LoRU-PartitionTop} & \cellcolor{gain!25}\textbf{49.61} & \cellcolor{gain!37}\textbf{14.47} & \cellcolor{gain!31}\underline{-0.63} & \cellcolor{gain!22}\textbf{79.22} & \cellcolor{gain!24}\textbf{92.97} & \cellcolor{gain!29}\textbf{71.81} & 53.95 & 66.77 & -24.58 & 60.14 & 80.23 & 0.33 & \underline{58.67} & \underline{25.30} & \underline{18.97} \\
\texttt{LoRU-WeightedAgg} & \cellcolor{gain!12}\underline{48.39} & \cellcolor{gain!36}\underline{14.20} & \cellcolor{gain!37}\textbf{1.23} & 70.78 & 88.92 & 55.26 & \underline{56.58} & \underline{70.50} & \underline{-1.56} & \underline{76.09} & \underline{84.23} & \underline{20.66} & 58.58 & 23.60 & 15.19 \\
\bottomrule
\end{tabular}%
}
\vspace{-0.5em}
\caption{Native estimators and their LoRU plug-ins ($\times100$). \textbf{Bold}/\underline{underline}: best/second-best per block; blue: gain over the native score (darker = larger).}
\vspace{-1em}
\label{tab:main_results}
\end{table*}

%% file: tables/loru_commitment.tex
\begin{table}[t]
\centering
\footnotesize
\setlength{\tabcolsep}{3.3pt}
\begin{tabular}{@{}lrrrr@{}}
\toprule
Model & Mentions & $H/\widetilde H$ & Mean $r(H)$ & $r\!\geq\!.9$ \\
\midrule
Llama-8B & 473 & $1.8\times$ & 0.60 & 10\% \\
Qwen-4B & 678 & $2.4\times$ & 0.64 & 29\% \\
Qwen-27B & 1,023 & $3.4\times$ & 0.64 & 29\% \\
\bottomrule
\end{tabular}
\caption{MuSiQue hop-answer mentions (first token). $H/\widetilde H$: median ratio to response-median entropy; $r(H)$: local rank of entropy, with the share $\geq0.9$ in the last column. Gold labels are used only here.}
\label{tab:loru_commitment}
\end{table}

%% file: tables/controlled_comp.tex
\begin{table}[t]
\centering
\footnotesize
\setlength{\tabcolsep}{3pt}
\begin{tabular}{@{}lrrrr@{}}
\toprule
$\Delta$AUROC & AIME & MuSiQue & DSQA & All \\
\midrule
$K$ vs.\ 5 tokens & $+18.81^{*}$ & $-7.85^{*}$ & $+5.23$ & $+5.39^{*}$ \\
Partition vs.\ global & $+0.22$ & $+3.25^{*}$ & $+2.48^{*}$ & $+1.98^{*}$ \\
\bottomrule
\end{tabular}
\caption{TokUR-EU selection controls, mean of three models ($^{*}$: 95\% CI excludes 0). $K$: PartitionTop's token count.}
\vspace{-1em}
\label{tab:loru_selection_controls}
\end{table}

%% file: sections/appendix_new.tex
\appendix

\section*{Appendix Overview}
\label{app:overview}

The appendix follows the order of the main text and is organized as follows.

\begin{itemize}
  \item \textbf{Appendix~\ref{app:propagation_controls}: Additional Analyses of Uncertainty Dependence} (Section~\ref{sec:uncertainty_propagation}). Controls for the natural-generation association, first-hop isolated uncertainty and the context-dependent confidence shift, and hop-distance correlations in natural generation (Table~\ref{tab:decay-hop}).
  \item \textbf{Appendix~\ref{app:loru_implementation}: LoRU Implementation Details} (Section~\ref{sec:relative_uncertainty}). Window eligibility, selection boundaries, sentence-anchored ranking, and computational requirements.
  \item \textbf{Appendix~\ref{app:datasets}: Dataset Details} (Section~\ref{sec:experimental_setup}). Benchmark statistics, subset construction, and answer extraction and correctness criteria (Table~\ref{tab:dataset_details}).
  \item \textbf{Appendix~\ref{app:experiment_details}: Experimental Configuration Details} (Section~\ref{sec:experimental_setup}). Uncertainty estimation, reasoning-step segmentation, ranking and scoring signals, operator-matched integration, and hyperparameters.
  \item \textbf{Appendix~\ref{app:aggregation_analysis}: Supplementary Aggregation Analysis} (Sections~\ref{sec:overall_effectiveness} and~\ref{sec:strategy_analysis}). Additional detection results, uncertainty at factual commitments, selection budget and evidence coverage, and ranking--scoring alignment.
  \item \textbf{Appendix~\ref{app:ablation}: Additional Ablation Results} (Section~\ref{sec:strategy_analysis}). Full locality--ranking ablations for TokUR-EU, MUR-Spike, and Mean Entropy, with benchmark-level breakdowns (Tables~\ref{tab:ablation_tokur}--\ref{tab:ablation_by_dataset}).
\end{itemize}

\section{Additional Analyses of Uncertainty Dependence}
\label{app:propagation_controls}

This section supplements the analyses of uncertainty dependence in Section~\ref{sec:uncertainty_propagation}.

\subsection{Controls for Natural Generation}
\label{app:natural_controls}

The in-context association between adjacent hops ($r=0.456$; Section~\ref{sec:propagation}) persists after accounting for hop position, hop length, and isolated hop uncertainty. Controlling for hop index and the log token lengths of both hops yields a partial correlation of $0.458$ ($[0.352,0.513]$), and additionally controlling for the isolated uncertainty of both hops yields $0.459$ ($[0.340,0.513]$). The isolated uncertainties of adjacent hops are also strongly correlated ($r=0.656$, $[0.492,0.759]$), but the previous hop's isolated uncertainty shows little association with the current hop's in-context uncertainty ($r=0.048$, $[-0.158,0.250]$). Part of the in-context association reflects response-level variation: after subtracting each response's mean uncertainty, the adjacent-hop correlation decreases to $0.171$ but remains positive (Table~\ref{tab:decay-hop}, $d=1$). Confidence intervals use 2{,}000 question-level bootstrap resamples (one response per question).

\subsection{First-Hop Uncertainty and the Context-Dependent Shift}
\label{app:first_hop_prefix}

In the controlled synthetic intervention, we examine whether the context-dependent confidence shift $\Delta$ (Eq.~\ref{eq:prefix_effect}) is associated with first-hop isolated uncertainty, measured as the negative mean token log-probability of $\tilde{a}_1$ given $q_1$. Across the 1{,}042 chains, this uncertainty shows little correlation with $\Delta$ (Pearson $r=0.023$, $[-0.039,0.085]$), including after controlling for the downstream answer's uncertainty under the isolated prompt (partial $r=0.008$, $[-0.053,0.072]$). The correlations are similarly small within each condition (Gold: $0.063$, $[-0.048,0.166]$; Close: $0.008$, $[-0.090,0.111]$; Far: $0.018$, $[-0.076,0.113]$). Across quartiles of first-hop isolated uncertainty, mean $\Delta$ ranges from $+0.23$ to $+0.37$ without a monotonic trend, and in-context confidence increases in 61--65\% of chains in each quartile. Confidence intervals use 2{,}000 question-level bootstrap resamples. Thus, these comparisons do not show a systematic association between first-hop isolated uncertainty and the observed confidence shift.

\subsection{Hop-Distance Correlation in Natural Generation}
\label{app:hop_decay}

Table~\ref{tab:decay-hop} extends the natural-generation analysis in Section~\ref{sec:propagation} from adjacent hops ($d=1$) to hop distances up to $d=6$. The correlation of in-context uncertainty decreases from $0.456$ at $d=1$ to $0.283$ at $d=6$. Restricting the analysis to the 41 chains in which every distance is observable yields closely matching values, suggesting that the decrease is not solely attributable to changing chain coverage at larger distances. After subtracting each response's mean uncertainty, correlations are substantially smaller, and their confidence intervals include zero from $d=3$ onward, although within-response centering can bias correlations downward for short chains. This hop-level pattern is consistent with the token-level decay in Section~\ref{sec:decay}.

\input{tables/decay_hop}

\section{LoRU Implementation Details}
\label{app:loru_implementation}

This section specifies boundary conditions and implementation details for the LoRU strategies defined in Section~\ref{sec:rank_aggregation}.

\subsection{Token-Level Computation}
\label{app:loru_token}

\paragraph{Window Eligibility.}
The sliding window $\mathcal{R}_t$ is defined only for $t>W$. Thus, the first $W$ tokens have no token-level local ranks and are excluded from rank-based token selection. All windows contain only preceding tokens, preventing future-token information from entering local comparisons.

\paragraph{Selection Boundaries.}
ScanTop considers candidate spans starting at positions $b>W$ and requires $T\geq W+B$. When no eligible span exists, its token-mean formulation falls back to the original full-response mean. PartitionTop divides positions $W+1,\ldots,T$ into non-overlapping partitions of length $L$, allowing a shorter final partition. Its score is defined when at least one eligible partition exists ($T>W$). Both strategies resolve ties by selecting the earliest eligible position or span, as specified by the minimum-argmax operators.

\paragraph{Sentence-Anchored Ranking.}
WeightedAgg fixes the comparison window at the beginning of each sentence rather than advancing it token by token. All tokens within a sentence are therefore ranked against the same preceding uncertainty values. Sentences starting within the first $W$ tokens receive a neutral weight $w_j=1/2$, ensuring that sentences without a complete preceding window can still contribute to the response score.

\paragraph{Computational Requirements.}
Given token-level ranking signals $a_t$ and scoring magnitudes $u_t$, local ranks can be computed through direct sliding-window comparisons in $O(TW)$ time. Subsequent selection or weighting operates on these precomputed values. LoRU therefore adds uncertainty aggregation operations without requiring additional language-model forward passes.

\subsection{Integration with Baseline Estimators}
\label{app:loru_integration}

Table~\ref{tab:loru_integration} summarizes how each LoRU strategy is applied to each baseline. All variants are computed on the same generated responses and token-level signals as the native estimator. ScanTop and PartitionTop rank the unperturbed predictive entropy, $a_t=H_t$, whereas WeightedAgg ranks the estimator's own token signal, $a_t=u_t$. In every case, $u_t$ supplies the uncertainty magnitudes used for scoring.

\begin{table*}[t]
\centering
\small
\setlength{\tabcolsep}{5pt}
\renewcommand{\arraystretch}{1.3}
\begin{tabular}{@{}lllllll@{}}
\toprule
Baseline & $u_t$ & Unit & Native score & ScanTop & PartitionTop & WeightedAgg \\
\midrule
TokUR-EU-Mean & $\mathrm{EU}_t$ & sentence & $\frac{1}{T}\sum_{t}u_t$ & Eq.~\ref{eq:scantop_score} & Eq.~\ref{eq:partitiontop_score} & Eq.~\ref{eq:weightedagg_score} \\
LL & $\ell_t$ & sentence & $\frac{1}{T}\sum_{t}u_t$ & Eq.~\ref{eq:scantop_score} & Eq.~\ref{eq:partitiontop_score} & Eq.~\ref{eq:weightedagg_score} \\
MUR-Spike & $\ell_t$ & step & $\max_{s\geq2}\delta_s$ & $\delta_{s(b^*)}$ & $\max_{s\geq2}\tilde{\delta}_s$ & -- \\
MUR-Momentum & $\ell_t$ & step & $\hat{M}_n$ & -- & $\hat{\tilde{M}}_n$ & Eq.~\ref{eq:mur_weighted} \\
Avg.\ Uncertainty & $\ell_t$ & step & $A_n$ & -- & $\tilde{A}_n$ & Eq.~\ref{eq:mur_weighted}, $\alpha=1$ \\
\midrule
\multicolumn{4}{@{}l}{Ranking signal $a_t$} & $H_t$ & $H_t$ & $u_t$ \\
\bottomrule
\end{tabular}
\caption{LoRU integration with each baseline estimator. $\mathrm{EU}_t$: TokUR token epistemic uncertainty; $\ell_t$: token negative log-likelihood; $H_t$: unperturbed predictive entropy. ``Unit'' is the segment used by WeightedAgg and by the step-level estimators. Step-level quantities are defined in Appendix~\ref{app:loru_integration}; tildes denote quantities recomputed from PartitionTop's within-step selections. --: combination not evaluated.}
\label{tab:loru_integration}
\end{table*}

\paragraph{Token-Mean Estimators.}
TokUR-EU-Mean and LL score a response by the mean of a token signal: $u_t=\mathrm{EU}_t$ for TokUR (Appendix~\ref{app:experiment_details}) and $u_t=\ell_t=-\log p(y_t\mid x,y_{<t})$ for LL. LoRU replaces this response-wide mean with Eqs.~\ref{eq:scantop_score}, \ref{eq:partitiontop_score}, and~\ref{eq:weightedagg_score}. ScanTop and PartitionTop select positions by the local ranks of $H_t$ and average $u_t$ at the selected positions. WeightedAgg computes sentence weights from the local ranks of $u_t$ and applies them to $u_t$.

\paragraph{Step-Level Notation.}
MUR and Avg.\ Uncertainty operate on reasoning steps $\mathcal{S}_1,\ldots,\mathcal{S}_n$ (Appendix~\ref{app:experiment_details}). The uncertainty of step $s$ is the mean token negative log-likelihood $m_s=|\mathcal{S}_s|^{-1}\sum_{t\in\mathcal{S}_s}\ell_t$. MUR tracks the momentum $M_s=\alpha M_{s-1}+(1-\alpha)m_s$, with $M_0=0$ and $\alpha=0.9$, and the step-wise uncertainty increase $\delta_s=m_s-M_{s-1}$. MUR-Spike scores a response by the largest increase, $\max_{s\geq2}\delta_s$, and MUR-Momentum by the bias-corrected final momentum $\hat{M}_n=M_n/(1-\alpha^n)$. Avg.\ Uncertainty replaces the momentum with the arithmetic average $A_s=s^{-1}\sum_{i\leq s}m_i$ and scores a response by $A_n$.

\paragraph{Step Uncertainty from PartitionTop.}
PartitionTop replaces each step uncertainty $m_s$ with a rank-selected estimate $\tilde{m}_s$. Within each step, positions $t>W$ are divided into consecutive partitions of length $L$, starting at the step's first eligible position. The token with the highest local rank of $H_t$ is selected from each partition, and $\tilde{m}_s$ is the mean of $\ell_t$ over the selected tokens. A step without eligible positions keeps $\tilde{m}_s=m_s$. Each step-level score is then recomputed on $\tilde{m}_s$: MUR-Spike uses $\max_{s\geq2}\tilde{\delta}_s$ with $\tilde{\delta}_s=\tilde{m}_s-\tilde{M}_{s-1}$, MUR-Momentum uses $\hat{\tilde{M}}_n=\tilde{M}_n/(1-\alpha^n)$, and Avg.\ Uncertainty uses $\tilde{A}_n=n^{-1}\sum_{s}\tilde{m}_s$, where $\tilde{M}_s$ follows the same momentum recursion as $M_s$.

\paragraph{Step Selection with ScanTop.}
MUR-Spike selects the step with the largest momentum deviation. LoRU-ScanTop replaces this selection with local relative ranking: it identifies the span $I_{b^*}$ with the highest average local rank of $H_t$ (Eq.~\ref{eq:scantop}) and scores the response by $\delta_{s(b^*)}$, the uncertainty increase of the step containing the span's first token. ScanTop is paired with MUR-Spike because both score a response by a single step.

\paragraph{Step Weighting with WeightedAgg.}
WeightedAgg applies the anchored ranking of Eq.~\ref{eq:sentence_anchor_rank} to reasoning steps instead of sentences. Each step receives a weight $\rho_s$, the mean local rank of $\ell_t$ over $\mathcal{S}_s$ against the window preceding the step's first token, with $\rho_s=1/2$ if the step starts within the first $W$ tokens. Since $\hat{M}_n=\sum_{s}\alpha^{n-s}m_s/\sum_{s}\alpha^{n-s}$, MUR-Momentum averages step uncertainties with exponentially decaying weights. WeightedAgg multiplies each of these weights by the step's local rank and token count:
\begin{equation}
    S_{\mathrm{WeightedAgg}}^{\mathrm{MUR}}(x,y)=
    \frac{\sum_{s=1}^{n}\alpha^{n-s}\rho_s|\mathcal{S}_s|\,m_s}
    {\sum_{s=1}^{n}\alpha^{n-s}\rho_s|\mathcal{S}_s|}.
    \label{eq:mur_weighted}
\end{equation}
For Avg.\ Uncertainty, we set $\alpha=1$, which recovers Eq.~\ref{eq:weightedagg_score} with reasoning steps as segments and $a_t=\ell_t$. WeightedAgg is paired with MUR-Momentum and Avg.\ Uncertainty because both average uncertainty over all steps.

\section{Dataset Details}
\label{app:datasets}

Table~\ref{tab:dataset_details} summarizes the three benchmarks used in Section~\ref{sec:experiments}. Prompts contain the question and answer-format instructions, without few-shot examples, supporting passages, or retrieved documents.

\begin{table*}[h]
\centering
\small
\setlength{\tabcolsep}{3pt}
\begin{tabular}{lccc}
\toprule
 & \textbf{AIME24--26} & \textbf{MuSiQue-4hop} & \textbf{DeepSearchQA} \\
\midrule
Domain              & Competition math & Multi-hop factual QA & Information seeking \\
Source              & AIME 2024--2026 & MuSiQue-Ans v1.0 (train) & Single Answer subset \\
\#Questions         & 90 & 458 & 316 \\
Avg.\ question length & 70.8 & 26.4 & 50.8 \\
Answer format       & Integer (0--999) & Short span (2.3) & Short span (2.2) \\
Gold answers / question & 1 & 1.44 & 1 \\
\bottomrule
\end{tabular}
\caption{Dataset statistics. Lengths are in words; the number in parentheses is the average gold answer length. MuSiQue-4hop gold answers include the aliases provided by the dataset.}
\label{tab:dataset_details}
\end{table*}

\paragraph{AIME24--26.}
We collect all 90 problems from the 2024, 2025, and 2026 American Invitational Mathematics Examination (AIME~I and II, 15 problems each, 30 per year). Every answer is an integer from 0 to 999. Three problems contain figures; we keep their Asymptote source in the problem text and do not render images.

\paragraph{MuSiQue-4hop.}
MuSiQue composes single-hop questions into multi-hop ones, so that each hop depends on the answer of the previous hop. We take the linear four-hop questions (type \texttt{4hop1}), in which the four hops form a single chain, from the MuSiQue-Ans v1.0 training split. Since no model is trained in our experiments, we draw from the training split, which contains more linear four-hop questions than the development split (648 vs.\ 246). We randomly sample 500 questions and exclude 42 for answer extractability or exact-match grading ambiguity: a preliminary run produced no final answer, predicted answers differed in granularity from the gold answer (e.g., a city versus its district), or the gold answer was time-sensitive or admitted multiple valid interpretations. The resulting 458 questions are used consistently in native-estimator and LoRU comparisons. We discard the supporting and distractor paragraphs that MuSiQue provides and keep only the question. Gold answers include the dataset's answer aliases; 113 questions have more than one acceptable answer.

\paragraph{DeepSearchQA.}
DeepSearchQA contains 900 information-seeking questions, each labeled with a problem category and an answer type. We keep the 316 questions whose answer type is \emph{Single Answer} and drop the 584 \emph{Set Answer} questions, whose answers are lists of entities and cannot be graded against a single final answer. The retained questions cover 14 categories, led by Finance \& Economics (56), Education (38), Politics \& Government (37), and Health (34). Although the benchmark was designed for agents with web search, we answer its questions closed-book, without retrieval or tool use.

\paragraph{Answer extraction and correctness.}
\begin{itemize}
  \item We extract the final answer from the answer line requested in the prompt; if the response ends with a \verb|\boxed{}| expression, we use its content instead. Intermediate boxed expressions are ignored.
  \item For AIME24--26, the extracted answer must be an integer with at most three digits and is compared numerically with the gold answer (e.g., \texttt{025} equals \texttt{25}).
  \item For the two question-answering datasets, we lowercase the prediction and the gold answers, remove punctuation and the articles \emph{a}, \emph{an}, and \emph{the}, and collapse whitespace. For DeepSearchQA, we additionally strip accents and Markdown emphasis. A response is correct if its normalized answer equals any normalized gold answer.
  \item Responses without an extractable final answer, or that reach the generation budget, are excluded from evaluation.
\end{itemize}

\section{Experimental Configuration Details}
\label{app:experiment_details}

This section provides configuration details complementing the experimental setup in Section~\ref{sec:experimental_setup}.

\paragraph{Uncertainty Estimation.}
For TokUR, we use the original model and nine perturbed copies. Gaussian perturbations with standard deviation $0.1$ are applied to the attention query and value projections along the leading eight left singular directions. For the hybrid Qwen architecture, only full-attention layers are perturbed. The MUR momentum coefficient is $0.9$ for configurations using its original momentum weighting.

\paragraph{Reasoning-Step Segmentation.}
For step-level estimators, Qwen responses are segmented using numbered step headers. Llama responses are segmented by paragraphs on AIME24--26 and DeepSearchQA, and by sentences on MuSiQue-4hop. Native estimators and their LoRU variants use identical response segmentation.

\paragraph{Ranking and Scoring Signals.}
ScanTop and PartitionTop compute local relative ranks from the unperturbed model's predictive entropy, while WeightedAgg uses the corresponding estimator's uncertainty signal. The original uncertainty magnitudes $u_t$ are retained for risk scoring, while relative ranks are computed from $a_t$. These signals are obtained from computations already required by the corresponding estimator, without additional model forward passes.

\paragraph{Operator-Matched Integration.}
For token-mean estimators, the three LoRU strategies directly modify response-level aggregation. For step-level estimators, LoRU modifies the aggregation operation associated with each strategy while preserving the remaining scoring structure. PartitionTop modifies within-step token aggregation; ScanTop modifies prominent-step selection in MUR-Spike; and WeightedAgg modifies cross-step weighting in MUR-Momentum and Avg. Uncertainty.

These combinations align each evaluated LoRU strategy with a corresponding aggregation operation in the native estimator.

\paragraph{Hyperparameters and Evaluation.}
The sliding-window size $W=20$, ScanTop span length $B=5$, and PartitionTop partition length $L=20$ are selected on a held-out validation set and fixed across experimental settings.

We use $2{,}000$ question-level bootstrap resamples for paired method comparisons. Each comparison evaluates native and LoRU-based scores on the same generated responses and correctness labels. Boundary handling for sliding windows and rank-based selection is described in Appendix~\ref{app:loru_implementation}.

\section{Supplementary Aggregation Analysis}
\label{app:aggregation_analysis}

This section provides additional results and details supporting Sections~\ref{sec:overall_effectiveness} and~\ref{sec:strategy_analysis}.

\paragraph{Additional Detection Results.}
On MuSiQue-4hop, LoRU-PartitionTop improves PRR in 13 of 15 model--estimator comparisons, complementing its AUROC improvements reported in Section~\ref{sec:overall_effectiveness}.

Improvements also occur across other reasoning tasks. On AIME24, integrating ScanTop into MUR-Spike increases AUROC from $62.22$ to $68.89$ on Llama-8B, from $63.57$ to $70.00$ on Qwen-4B, and from $55.80$ to $65.22$ on Qwen-27B. On MuSiQue-4hop, ScanTop improves TokUR-EU-Mean from $37.65$ to $54.64$ AUROC on Qwen-27B. WeightedAgg also improves Avg. Uncertainty from $80.71$ to $90.71$ AUROC on Qwen-4B AIME24.

These examples show improvements across multiple models, tasks, and aggregation strategies, with effects varying across configurations.

\paragraph{Uncertainty at Factual Commitments.}
The analysis in Table~\ref{tab:loru_commitment} identifies the first generated mentions of matched MuSiQue hop sub-answers, excluding copies from the question. These positions have elevated predictive entropy relative to the response median, with local relative ranks reflecting their uncertainty compared with preceding tokens.

Answer types differ between approximately $68\%$ of adjacent MuSiQue hops. In contrast, approximately $35$--$50\%$ of AIME tokens are digits or mathematical symbols, whose median predictive entropy ranges from $0.000$ to $0.004$. These differences provide additional context for the heterogeneous distribution of uncertainty evidence across reasoning tasks.

\paragraph{Selection Budget and Evidence Coverage.}
The controlled comparisons in Table~\ref{tab:loru_selection_controls} distinguish the amount of retained uncertainty evidence from its distribution across the response. Increasing retention from five tokens to the partition-wise selection budget yields an average AUROC gain of $5.4$ points across nine evaluated settings, although the effects vary by benchmark ($+18.81$ on AIME and $-7.85$ on MuSiQue).

After matching the number of selected tokens and the selection unit, partition-wise selection retains a $2.0$-point AUROC advantage over global selection, with a $95\%$ bootstrap confidence interval of $[+0.3,+3.6]$. This comparison provides evidence that distributing selected tokens across the response can be beneficial beyond increasing the selection budget alone.

\paragraph{Ranking--Scoring Alignment.}
On MuSiQue TokUR-EU, ScanTop changes AUROC by $+8.16$ and $+16.99$ points on the two Qwen models, compared with $-3.08$ points on Llama.

The corresponding within-response Spearman correlations between predictive entropy, used as the ranking signal, and epistemic uncertainty, used for risk scoring, are $0.92$--$0.97$ on Qwen and $-0.14$ on Llama. This contrast is consistent with ranking--scoring alignment being relevant to concentrated evidence selection, but does not establish that alignment alone causes the observed performance differences.

These results complement the analysis in Section~\ref{sec:strategy_analysis}, showing that evidence coverage and ranking--scoring alignment are distinct considerations when choosing an aggregation strategy.

\section{Additional Ablation Results}
\label{app:ablation}

This section reports the full ablation results summarized in Section~\ref{sec:strategy_analysis}. We investigate the complementarity of locality and relative ranking through a $2\times2$ ablation of LoRU-ScanTop. The four configurations combine two comparison ranges (a local sliding window or the entire preceding prefix) with two ranking criteria (relative rank or z-score). We additionally compare selection using the raw uncertainty signal, random span selection, and the native estimator score.

All comparisons use the same generated responses and scoring signals. Results are averaged over nine evaluated model--dataset settings and scaled by $100$. We report AUROC, AUPRC, and PRR, with differences computed relative to the complete ScanTop configuration. Confidence intervals are obtained from $2{,}000$ paired question-level bootstrap resamples.

For a metric $M$, we define the interaction between locality and ranking as
\begin{equation}
\begin{aligned}
I_M ={}&
\bigl(M_{\mathrm{window,rank}}
      -M_{\mathrm{window,z}}\bigr)\\
&-\bigl(M_{\mathrm{prefix,rank}}
      -M_{\mathrm{prefix,z}}\bigr).
\end{aligned}
\label{eq:ablation_interaction}
\end{equation}
A positive interaction indicates that the advantage of relative ranking over z-scoring is greater within a local window than under full-prefix comparisons.

\paragraph{TokUR-EU Ablation.}
Table~\ref{tab:ablation_tokur} reports the full TokUR-EU results. Combining locality and relative ranking produces higher mean AUROC than either windowed z-scoring or full-prefix ranking, with a positive interaction of $+4.7$ points and a confidence interval excluding zero. However, individual component comparisons are not statistically conclusive, and the complete ScanTop configuration does not significantly outperform the full-prefix z-score or native TokUR-EU readout. Thus, the positive interaction supports complementarity in this ablation, rather than a consistent standalone advantage of either component.

\begin{table*}[t]
\centering
\scriptsize
\setlength{\tabcolsep}{3pt}
\resizebox{\textwidth}{!}{%
\begin{tabular}{@{}lrrrlll@{}}
\toprule
Configuration & AUROC & AUPRC & PRR
& $\Delta$AUROC [95\% CI]
& $\Delta$AUPRC [95\% CI]
& $\Delta$PRR [95\% CI] \\
\midrule
ScanTop (window + rank)
& 56.6 & 26.8 & 9.4
& -- & -- & -- \\
Window + z-score
& 54.6 & 26.0 & 6.3
& $-2.0\ [-5.1,+1.3]$
& $-0.8\ [-3.2,+1.7]$
& $-3.2\ [-9.5,+3.3]$ \\
Prefix + rank
& 53.5 & 28.3 & 4.9
& $-3.1\ [-10.0,+2.9]$
& $+1.5\ [-5.0,+9.4]$
& $-4.6\ [-17.6,+7.0]$ \\
Prefix + z-score
& 56.2 & 27.5 & 9.8
& $-0.5\ [-7.1,+5.5]$
& $+0.7\ [-3.4,+10.1]$
& $+0.4\ [-12.1,+12.3]$ \\
Raw signal
& 51.8 & 25.7 & 2.5
& $-4.8\ [-11.1,+2.1]$
& $-1.1\ [-5.3,+8.2]$
& $-6.9\ [-19.6,+5.6]$ \\
Random 5-token span
& 54.8 & 27.0 & 7.8
& $-1.8\ [-8.8,+4.3]$
& $+0.2\ [-4.0,+6.3]$
& $-1.6\ [-14.2,+10.3]$ \\
Native TokUR-EU
& 59.2 & 28.0 & 12.4
& $+2.6\ [-2.3,+7.3]$
& $+1.2\ [-2.1,+4.8]$
& $+2.9\ [-7.4,+12.8]$ \\
\midrule
Interaction
& -- & -- & --
& $\mathbf{+4.7}\ [+0.6,+8.8]$
& $0.0\ [-6.0,+5.5]$
& $+8.2\ [-0.5,+17.9]$ \\
\bottomrule
\end{tabular}%
}
\caption{TokUR-EU ScanTop ablation, averaged across nine model--dataset settings. Metric differences are relative to complete ScanTop; the interaction is defined in Eq.~\ref{eq:ablation_interaction}. Confidence intervals use paired question-level bootstrap resampling.}
\label{tab:ablation_tokur}
\end{table*}

\paragraph{Cross-Estimator Differences.}
We repeat the ablation with Mean Entropy and MUR-Spike to examine whether the same pattern appears across uncertainty estimators.

For MUR-Spike, all tested simplifications yield lower mean AUROC than complete ScanTop, but the individual differences and interaction are not statistically conclusive (Table~\ref{tab:ablation_mur}).

Mean Entropy exhibits a different pattern (Table~\ref{tab:ablation_entropy}). Full-prefix ranking, full-prefix z-scoring, and raw-signal selection outperform local-window ScanTop on average, with confidence intervals excluding zero. Although its interaction is positive, these results show that the benefit of local-window ranking depends on the uncertainty estimator and is not universal.

\begin{table*}[t]
\centering
\scriptsize
\setlength{\tabcolsep}{3pt}
\resizebox{\textwidth}{!}{%
\begin{tabular}{@{}lrrrlll@{}}
\toprule
Configuration & AUROC & AUPRC & PRR
& $\Delta$AUROC [95\% CI]
& $\Delta$AUPRC [95\% CI]
& $\Delta$PRR [95\% CI] \\
\midrule
ScanTop (window + rank)
& 56.7 & 27.6 & 11.0
& -- & -- & -- \\
Window + z-score
& 56.0 & 27.2 & 10.6
& $-0.7\ [-3.7,+2.3]$
& $-0.4\ [-2.0,+5.0]$
& $-0.4\ [-5.9,+6.0]$ \\
Prefix + rank
& 53.5 & 26.9 & 6.6
& $-3.1\ [-7.9,+1.7]$
& $-0.7\ [-3.0,+1.8]$
& $-4.4\ [-11.2,+2.5]$ \\
Prefix + z-score
& 55.7 & 26.8 & 8.8
& $-1.0\ [-5.6,+3.6]$
& $-0.8\ [-2.9,+2.2]$
& $-2.2\ [-9.8,+5.2]$ \\
Raw signal
& 52.7 & 25.6 & 5.4
& $-4.0\ [-9.7,+2.4]$
& $-2.0\ [-4.1,+0.9]$
& $-5.6\ [-13.7,+2.9]$ \\
Native MUR-Spike
& 53.9 & 25.7 & 2.0
& $-2.8\ [-7.9,+2.4]$
& $-1.9\ [-4.3,+0.8]$
& $-9.0\ [-17.7,+0.3]$ \\
\midrule
Interaction
& -- & -- & --
& $+2.8\ [-2.2,+7.8]$
& $+0.4\ [-5.2,+3.2]$
& $+2.6\ [-6.4,+10.9]$ \\
\bottomrule
\end{tabular}%
}
\caption{MUR-Spike ScanTop ablation across nine model--dataset settings. Differences are relative to complete ScanTop.}
\label{tab:ablation_mur}
\end{table*}

\begin{table*}[t]
\centering
\scriptsize
\setlength{\tabcolsep}{3pt}
\resizebox{\textwidth}{!}{%
\begin{tabular}{@{}lrrrlll@{}}
\toprule
Configuration & AUROC & AUPRC & PRR
& $\Delta$AUROC [95\% CI]
& $\Delta$AUPRC [95\% CI]
& $\Delta$PRR [95\% CI] \\
\midrule
ScanTop (window + rank)
& 53.9 & 25.2 & 4.0
& -- & -- & -- \\
Window + z-score
& 51.7 & 24.6 & 1.0
& $-2.2\ [-6.4,+1.8]$
& $-0.6\ [-2.5,+3.9]$
& $-3.0\ [-9.1,+3.4]$ \\
Prefix + rank
& 59.3 & 29.2 & 16.2
& $+5.3\ [+0.4,+10.3]$
& $+4.0\ [+1.3,+9.7]$
& $+12.1\ [+3.6,+20.3]$ \\
Prefix + z-score
& 62.3 & 32.7 & 20.1
& $+8.4\ [+4.0,+12.5]$
& $+7.5\ [+2.1,+13.8]$
& $+16.1\ [+7.2,+24.3]$ \\
Raw signal
& 62.6 & 32.0 & 23.3
& $+8.7\ [+4.2,+13.3]$
& $+6.8\ [+3.6,+14.2]$
& $+19.3\ [+10.7,+27.5]$ \\
Random 5-token span
& 53.9 & 26.7 & 6.5
& $+0.0\ [-7.8,+7.4]$
& $+1.5\ [-2.5,+7.6]$
& $+2.5\ [-10.4,+14.8]$ \\
Native Mean Entropy
& 61.9 & 28.5 & 13.5
& $+8.0\ [+3.2,+12.9]$
& $+3.3\ [+0.3,+5.9]$
& $+9.5\ [-0.2,+19.0]$ \\
\midrule
Interaction
& -- & -- & --
& $\mathbf{+5.2}\ [+1.0,+9.5]$
& $+4.1\ [-1.1,+9.1]$
& $+7.0\ [-0.9,+14.0]$ \\
\bottomrule
\end{tabular}%
}
\caption{Mean Entropy ScanTop ablation across nine model--dataset settings. Differences are relative to complete ScanTop. Mean Entropy is included as an additional ablation diagnostic.}
\label{tab:ablation_entropy}
\end{table*}

\paragraph{Benchmark-Level Results.}
Table~\ref{tab:ablation_by_dataset} further breaks down AUROC differences by benchmark. For TokUR-EU, replacing local ranks with windowed z-scores particularly reduces performance on MuSiQue, while the positive interaction is most apparent on MuSiQue and DeepSearchQA. Mean Entropy shows stronger performance from full-prefix comparisons, especially on AIME. MUR exhibits smaller and less consistent differences across datasets.

\begin{table*}[t]
\centering
\scriptsize
\setlength{\tabcolsep}{4pt}
\resizebox{\textwidth}{!}{%
\begin{tabular}{@{}llccc@{}}
\toprule
Estimator & Configuration
& AIME24--26 & MuSiQue-4hop & DeepSearchQA \\
\midrule
\multicolumn{5}{l}{\textit{TokUR-EU}}\\
& Window + z-score
& $+5.0\ [-1.7,+11.9]$
& $-6.6\ [-10.7,-2.5]$
& $-4.5\ [-10.0,+0.8]$ \\
& Prefix + rank
& $-2.8\ [-21.7,+12.8]$
& $-1.9\ [-6.7,+3.3]$
& $-4.6\ [-11.5,+2.3]$ \\
& Prefix + z-score
& $+1.9\ [-16.7,+16.7]$
& $-2.4\ [-7.8,+3.2]$
& $-0.9\ [-8.4,+6.1]$ \\
& Raw signal
& $-9.2\ [-25.8,+8.3]$
& $-2.7\ [-8.8,+3.5]$
& $-2.4\ [-9.0,+4.0]$ \\
& Random 5-token span
& $-3.3\ [-22.2,+12.0]$
& $-2.0\ [-9.3,+5.8]$
& $-0.1\ [-9.2,+8.4]$ \\
& Native readout
& $+10.6\ [-1.6,+21.7]$
& $-7.4\ [-13.0,-1.1]$
& $+4.5\ [-1.6,+10.0]$ \\
& Interaction
& $-0.3\ [-9.2,+8.4]$
& $+6.0\ [+0.0,+11.7]$
& $+8.2\ [+1.8,+14.8]$ \\
\midrule
\multicolumn{5}{l}{\textit{MUR-Spike}}\\
& Window + z-score
& $+3.2\ [-3.0,+9.1]$
& $-1.7\ [-5.8,+2.6]$
& $-3.6\ [-8.9,+1.7]$ \\
& Prefix + rank
& $-11.4\ [-23.2,+1.0]$
& $+1.8\ [-3.5,+6.7]$
& $+0.2\ [-6.4,+6.9]$ \\
& Prefix + z-score
& $-3.3\ [-14.1,+7.4]$
& $+0.0\ [-6.2,+6.1]$
& $+0.2\ [-6.3,+7.0]$ \\
& Raw signal
& $-9.8\ [-23.2,+7.9]$
& $-1.6\ [-7.8,+4.4]$
& $-0.5\ [-7.2,+6.5]$ \\
& Native readout
& $-11.5\ [-22.9,+0.3]$
& $+2.6\ [-4.6,+9.3]$
& $+0.5\ [-7.0,+7.5]$ \\
& Interaction
& $+4.9\ [-8.1,+17.5]$
& $0.0\ [-5.3,+5.5]$
& $+3.6\ [-2.8,+10.2]$ \\
\midrule
\multicolumn{5}{l}{\textit{Mean Entropy}}\\
& Window + z-score
& $-1.8\ [-13.0,+6.7]$
& $-0.4\ [-4.9,+4.3]$
& $-4.4\ [-10.8,+2.4]$ \\
& Prefix + rank
& $+12.8\ [+1.9,+23.6]$
& $+3.5\ [-2.9,+9.5]$
& $-0.3\ [-7.2,+6.2]$ \\
& Prefix + z-score
& $+16.9\ [+8.0,+25.5]$
& $+5.0\ [-1.5,+11.0]$
& $+3.1\ [-3.8,+9.8]$ \\
& Raw signal
& $+17.4\ [+7.5,+26.8]$
& $+5.4\ [-0.9,+11.5]$
& $+3.3\ [-3.9,+10.2]$ \\
& Random 5-token span
& $-0.2\ [-19.6,+17.7]$
& $-2.1\ [-10.4,+6.5]$
& $+2.4\ [-7.2,+12.0]$ \\
& Native readout
& $+15.8\ [+4.0,+27.6]$
& $-2.2\ [-8.1,+4.0]$
& $+10.3\ [+4.8,+16.1]$ \\
& Interaction
& $+6.0\ [-2.8,+15.0]$
& $+1.9\ [-3.7,+7.1]$
& $+7.9\ [-0.1,+15.6]$ \\
\bottomrule
\end{tabular}%
}
\caption{Benchmark-level AUROC differences relative to complete LoRU-ScanTop, averaged across three models per benchmark, with 95\% paired bootstrap confidence intervals. Interaction values follow Eq.~\ref{eq:ablation_interaction}.}
\label{tab:ablation_by_dataset}
\end{table*}

%% file: tables/decay_hop.tex
\begin{table*}[t]
\centering
\small
\setlength{\tabcolsep}{4pt}
\begin{tabular}{rrccc}
\toprule
$d$ & Pairs & All chains & $\geq$7 hops & Within-response \\
\midrule
1 & 839 & 0.456 [0.37, 0.51] & 0.473 [0.37, 0.54] & 0.171 [0.03, 0.28] \\
2 & 739 & 0.416 [0.35, 0.47] & 0.420 [0.35, 0.47] & 0.130 [0.02, 0.22] \\
3 & 644 & 0.349 [0.26, 0.41] & 0.363 [0.27, 0.43] & 0.051 [-0.11, 0.19] \\
4 & 566 & 0.346 [0.23, 0.43] & 0.342 [0.23, 0.42] & 0.069 [-0.09, 0.23] \\
5 & 503 & 0.296 [0.17, 0.37] & 0.301 [0.18, 0.37] & 0.010 [-0.16, 0.16] \\
6 & 455 & 0.283 [0.15, 0.37] & 0.283 [0.15, 0.38] & -0.002 [-0.15, 0.14] \\
\bottomrule
\end{tabular}
\caption{Hop-distance correlation of in-context uncertainty on DeepSearchQA, with 95\% question-level bootstrap intervals. ``$\geq$7 hops'' restricts to the 41 chains in which every distance is observable. ``Within-response'' subtracts each response's mean uncertainty before computing the correlation; this removes the shared response-level component but biases correlations downward for short chains.}
\label{tab:decay-hop}
\end{table*}

%% file: acl_latex.bbl
\begin{thebibliography}{29}
\providecommand{\natexlab}[1]{#1}

\bibitem[{Amit et~al.(2026)Amit, Galil, and El-Yaniv}]{amit2026evaluating}
Ido Amit, Ido Galil, and Ran El-Yaniv. 2026.
\newblock Evaluating llm uncertainty in long-form generation using deterministic ground truth.
\newblock \emph{arXiv preprint arXiv:2607.03870}.

\bibitem[{Bakman et~al.(2025)Bakman, Yaldiz, Kang, Zhang, Buyukates, Avestimehr, and Karimireddy}]{bakman2025reconsidering}
Yavuz~Faruk Bakman, Duygu~Nur Yaldiz, Sungmin Kang, Tuo Zhang, Baturalp Buyukates, Salman Avestimehr, and Sai~Praneeth Karimireddy. 2025.
\newblock Reconsidering llm uncertainty estimation methods in the wild.
\newblock In \emph{Proceedings of the 63rd Annual Meeting of the Association for Computational Linguistics (Volume 1: Long Papers)}, pages 29531--29556.

\bibitem[{Chen et~al.(2025)Chen, Chen, Zhou, Tao, Ding, Xie, Xie, Li, and Feng}]{chen2025enhancing}
Kedi Chen, Qin Chen, Jie Zhou, Xinqi Tao, Bowen Ding, Jingwen Xie, Mingchen Xie, Peilong Li, and Zheng Feng. 2025.
\newblock Enhancing uncertainty modeling with semantic graph for hallucination detection.
\newblock In \emph{Proceedings of the AAAI Conference on Artificial Intelligence}, volume~39, pages 23586--23594.

\bibitem[{Chuang et~al.(2024)Chuang, Qiu, Hsieh, Krishna, Kim, and Glass}]{chuang2024lookback}
Yung-Sung Chuang, Linlu Qiu, Cheng-Yu Hsieh, Ranjay Krishna, Yoon Kim, and James Glass. 2024.
\newblock Lookback lens: Detecting and mitigating contextual hallucinations in large language models using only attention maps.
\newblock In \emph{Proceedings of the 2024 Conference on Empirical Methods in Natural Language Processing}, pages 1419--1436.

\bibitem[{Duan et~al.(2024)Duan, Cheng, Wang, Zavalny, Wang, Xu, Kailkhura, and Xu}]{duan2024shifting}
Jinhao Duan, Hao Cheng, Shiqi Wang, Alex Zavalny, Chenan Wang, Renjing Xu, Bhavya Kailkhura, and Kaidi Xu. 2024.
\newblock Shifting attention to relevance: Towards the predictive uncertainty quantification of free-form large language models.
\newblock In \emph{Proceedings of the 62nd Annual Meeting of the Association for Computational Linguistics (Volume 1: Long Papers)}, pages 5050--5063.

\bibitem[{Fadeeva et~al.(2024)Fadeeva, Rubashevskii, Shelmanov, Petrakov, Li, Mubarak, Tsymbalov, Kuzmin, Panchenko, Baldwin, Nakov, and Panov}]{fadeeva2024fact}
Ekaterina Fadeeva, Aleksandr Rubashevskii, Artem Shelmanov, Sergey Petrakov, Haonan Li, Hamdy Mubarak, Evgenii Tsymbalov, Gleb Kuzmin, Alexander Panchenko, Timothy Baldwin, Preslav Nakov, and Maxim Panov. 2024.
\newblock Fact-checking the output of large language models via token-level uncertainty quantification.
\newblock In \emph{In Findings of the Association for Computational Linguistics ACL 2024}, pages 9367--9385.

\bibitem[{Farquhar et~al.(2024)Farquhar, Kossen, Kuhn, and Gal}]{farquhar2024detecting}
Sebastian Farquhar, Jannik Kossen, Lorenz Kuhn, and Yarin Gal. 2024.
\newblock Detecting hallucinations in large language models using semantic entropy.
\newblock \emph{Nature}, 630(8017):625--630.

\bibitem[{Grattafiori et~al.(2024)Grattafiori, Dubey, Jauhri, Pandey, Kadian, Al-Dahle, Letman, Mathur, Schelten, Vaughan et~al.}]{grattafiori2024llama}
Aaron Grattafiori, Abhimanyu Dubey, Abhinav Jauhri, Abhinav Pandey, Abhishek Kadian, Ahmad Al-Dahle, Aiesha Letman, Akhil Mathur, Alan Schelten, Alex Vaughan, et~al. 2024.
\newblock The llama 3 herd of models.
\newblock \emph{arXiv preprint arXiv:2407.21783}.

\bibitem[{Gupta et~al.(2026)Gupta, Chatterjee, Haas, Tao, Wang, Liu, Oiwa, Gribovskaya, Ackermann, Blitzer et~al.}]{gupta2026deepsearchqa}
Nikita Gupta, Riju Chatterjee, Lukas Haas, Connie Tao, Andrew Wang, Chang Liu, Hidekazu Oiwa, Elena Gribovskaya, Jan Ackermann, John Blitzer, et~al. 2026.
\newblock Deepsearchqa: Bridging the comprehensiveness gap for deep research agents.
\newblock \emph{arXiv preprint arXiv:2601.20975}.

\bibitem[{Huang et~al.(2024)Huang, Li, Yu, Sesia, Hassani, Lee, Bastani, and Dobriban}]{huang2024uncertainty}
Xinmeng Huang, Shuo Li, Mengxin Yu, Matteo Sesia, Hamed Hassani, Insup Lee, Osbert Bastani, and Edgar Dobriban. 2024.
\newblock Uncertainty in language models: Assessment through rank-calibration.
\newblock In \emph{Proceedings of the 2024 Conference on Empirical Methods in Natural Language Processing}, pages 284--312.

\bibitem[{Kossen et~al.(2024)Kossen, Han, Razzak, Schut, Malik, and Gal}]{kossen2024semantic}
Jannik Kossen, Jiatong Han, Muhammed Razzak, Lisa Schut, Shreshth Malik, and Yarin Gal. 2024.
\newblock Semantic entropy probes: Robust and cheap hallucination detection in llms.
\newblock \emph{arXiv preprint arXiv:2406.15927}.

\bibitem[{Kuhn et~al.(2023)Kuhn, Gal, and Farquhar}]{kuhn2023semantic}
Lorenz Kuhn, Yarin Gal, and Sebastian Farquhar. 2023.
\newblock Semantic uncertainty: Linguistic invariances for uncertainty estimation in natural language generation.
\newblock \emph{arXiv preprint arXiv:2302.09664}.

\bibitem[{Li et~al.(2026)Li, Channing, Bhat, Jones, and Gal}]{li2026building}
Lin Li, Georgia Channing, Suhaas~M Bhat, Gabriel~Davis Jones, and Yarin Gal. 2026.
\newblock Building reliable long-form generation via hallucination rejection sampling.
\newblock \emph{arXiv preprint arXiv:2606.03628}.

\bibitem[{Manakul et~al.(2023)Manakul, Liusie, and Gales}]{manakul2023selfcheckgpt}
Potsawee Manakul, Adian Liusie, and Mark Gales. 2023.
\newblock Selfcheckgpt: Zero-resource black-box hallucination detection for generative large language models.
\newblock In \emph{Proceedings of the 2023 conference on empirical methods in natural language processing}, pages 9004--9017.

\bibitem[{Mao et~al.(2025)Mao, Bisliouk, Nama, and Ruchkin}]{mao2025temporalizing}
Zhenjiang Mao, Artem Bisliouk, Rohith Nama, and Ivan Ruchkin. 2025.
\newblock Temporalizing confidence: Evaluation of chain-of-thought reasoning with signal temporal logic.
\newblock In \emph{Proceedings of the 20th Workshop on Innovative Use of NLP for Building Educational Applications (BEA 2025)}, pages 882--890.

\bibitem[{Nikitin et~al.(2024)Nikitin, Kossen, Gal, and Marttinen}]{nikitin2024kernel}
Alexander Nikitin, Jannik Kossen, Yarin Gal, and Pekka Marttinen. 2024.
\newblock Kernel language entropy: Fine-grained uncertainty quantification for llms from semantic similarities.
\newblock \emph{Advances in Neural Information Processing Systems}, 37:8901--8929.

\bibitem[{NS et~al.()NS, Yin, and Li}]{nsgauss}
Karthik~Somayaji NS, Yuxuan Yin, and Peng Li.
\newblock Gauss: Graph-assisted uncertainty quantification using structure and semantics for long-form generation in llms.
\newblock In \emph{Forty-third International Conference on Machine Learning}.

\bibitem[{Qiao et~al.(2026)Qiao, Pan, Mi, Liu, Shen, Sun, and Chu}]{qiao2026lowest}
Yitong Qiao, Licheng Pan, Yu~Mi, Lei Liu, Yue Shen, Fei Sun, and Zhixuan Chu. 2026.
\newblock Lowest span confidence: A zero-shot metric for efficient and black-box hallucination detection in llms.
\newblock \emph{arXiv preprint arXiv:2601.19918}.

\bibitem[{Shelmanov et~al.(2025)Shelmanov, Fadeeva, Tsvigun, Tsvigun, Xie, Kiselev, Daheim, Zhang, Vazhentsev, Sachan et~al.}]{shelmanov2025head}
Artem Shelmanov, Ekaterina Fadeeva, Akim Tsvigun, Ivan Tsvigun, Zhuohan Xie, Igor Kiselev, Nico Daheim, Caiqi Zhang, Artem Vazhentsev, Mrinmaya Sachan, et~al. 2025.
\newblock A head to predict and a head to question: Pre-trained uncertainty quantification heads for hallucination detection in llm outputs.
\newblock In \emph{Proceedings of the 2025 Conference on Empirical Methods in Natural Language Processing}, pages 35700--35719.

\bibitem[{Trivedi et~al.(2022)Trivedi, Balasubramanian, Khot, and Sabharwal}]{trivedi2022musique}
Harsh Trivedi, Niranjan Balasubramanian, Tushar Khot, and Ashish Sabharwal. 2022.
\newblock musique: Multihop questions via single-hop question composition.
\newblock \emph{Transactions of the Association for Computational Linguistics}, 10:539--554.

\bibitem[{Vashurin et~al.(2025)Vashurin, Fadeeva, Vazhentsev, Rvanova, Vasilev, Tsvigun, Petrakov, Xing, Sadallah, Grishchenkov et~al.}]{vashurin2025benchmarking}
Roman Vashurin, Ekaterina Fadeeva, Artem Vazhentsev, Lyudmila Rvanova, Daniil Vasilev, Akim Tsvigun, Sergey Petrakov, Rui Xing, Abdelrahman Sadallah, Kirill Grishchenkov, et~al. 2025.
\newblock Benchmarking uncertainty quantification methods for large language models with lm-polygraph.
\newblock \emph{Transactions of the Association for Computational Linguistics}, 13:220--248.

\bibitem[{Vazhentsev et~al.(2025)Vazhentsev, Rvanova, Lazichny, Panchenko, Panov, Baldwin, and Shelmanov}]{vazhentsev2025token}
Artem Vazhentsev, Lyudmila Rvanova, Ivan Lazichny, Alexander Panchenko, Maxim Panov, Timothy Baldwin, and Artem Shelmanov. 2025.
\newblock Token-level density-based uncertainty quantification methods for eliciting truthfulness of large language models.
\newblock In \emph{Proceedings of the 2025 Conference of the Nations of the Americas Chapter of the Association for Computational Linguistics: Human Language Technologies (Volume 1: Long Papers)}, pages 2246--2262.

\bibitem[{Wang et~al.(2026)Wang, Liu, Wang, Huang, Shah, and Wei}]{wang2026inference}
Yu~Wang, Minghao Liu, Jiayun Wang, Jinrui Huang, Ankit Shah, and Wei Wei. 2026.
\newblock Inference time optimization with confidence dynamics.
\newblock \emph{arXiv preprint arXiv:2605.25244}.

\bibitem[{Xu et~al.(2026)Xu, He, Lu, Sun, Li, Lam, and Hao}]{xu2026unveiling}
Ting Xu, Xu~He, Yupu Lu, Jiankai Sun, Dong Li, Wai Lam, and Jianye Hao. 2026.
\newblock Unveiling the entropy dynamics of chain-of-thought reasoning.
\newblock \emph{arXiv preprint arXiv:2606.02020}.

\bibitem[{Yan et~al.(2026)Yan, Xu, Xu, Li, Zhang, Luo, Wu, Tuan, Zhao, Lin et~al.}]{yan2026mur}
Hang Yan, Fangzhi Xu, Rongman Xu, Yifei Li, Jian Zhang, Haoran Luo, Xiaobao Wu, Luu~Anh Tuan, Haiteng Zhao, Qika Lin, et~al. 2026.
\newblock Mur: Momentum uncertainty guided reasoning for large language models.
\newblock In \emph{Proceedings of the 64th Annual Meeting of the Association for Computational Linguistics (Volume 1: Long Papers)}, pages 23078--23103.

\bibitem[{Zhang et~al.(2023)Zhang, Qiu, Guo, Deng, Zhang, Zhang, Zhou, Wang, and Fu}]{zhang2023enhancing}
Tianhang Zhang, Lin Qiu, Qipeng Guo, Cheng Deng, Yue Zhang, Zheng Zhang, Chenghu Zhou, Xinbing Wang, and Luoyi Fu. 2023.
\newblock Enhancing uncertainty-based hallucination detection with stronger focus.
\newblock In \emph{Proceedings of the 2023 Conference on Empirical Methods in Natural Language Processing}, pages 915--932.

\bibitem[{Zhang et~al.(2026)Zhang, Shi, Wang, Wang, He, Li, Chen, Han, Xu, Zhang et~al.}]{zhang2026tokur}
Tunyu Zhang, Haizhou Shi, Yibin Wang, Hengyi Wang, Xiaoxiao He, Zhuowei Li, Haoxian Chen, Ligong Han, Kai Xu, Huan Zhang, et~al. 2026.
\newblock Tokur: Token-level uncertainty estimation for large language model reasoning.
\newblock In \emph{International Conference on Learning Representations}, volume 2026, pages 52910--52942.

\bibitem[{Zhao et~al.(2025)Zhao, Li, Liu, Cheng, Sun, Oishi, Osaki, Matsuda, Yao, Zhao et~al.}]{zhao2025uncertainty}
Qiwei Zhao, Dong Li, Yanchi Liu, Wei Cheng, Yiyou Sun, Mika Oishi, Takao Osaki, Katsushi Matsuda, Huaxiu Yao, Chen Zhao, et~al. 2025.
\newblock Uncertainty propagation on llm agent.
\newblock In \emph{Proceedings of the 63rd Annual Meeting of the Association for Computational Linguistics (Volume 1: Long Papers)}, pages 6064--6073.

\bibitem[{Zhu et~al.(2026)Zhu, Wu, Zeng, Xu, Kang, Guo, Lu, Huang, and Zhou}]{zhu2026edis}
Chenghua Zhu, Siyan Wu, Xiangkang Zeng, Zishan Xu, Zhaolu Kang, Yifu Guo, Yuquan Lu, Junduan Huang, and Guojing Zhou. 2026.
\newblock Edis: Diagnosing llm reasoning via entropy dynamics.
\newblock \emph{arXiv preprint arXiv:2602.01288}.

\end{thebibliography}
